\documentclass[10pt,twocolumn,letterpaper]{article}

\usepackage[pagenumbers]{cvpr} 

\usepackage{amsmath,amsfonts,bm}

\def\eqref#1{equation~\ref{#1}}

\def\1{\bm{1}}

\def\vc{{\bm{c}}}

\def\vm{{\bm{m}}}

\def\vx{{\bm{x}}}

\def\vz{{\bm{z}}}

\def\mM{{\bm{M}}}

\def\mZ{{\bm{Z}}}

\DeclareMathAlphabet{\mathsfit}{\encodingdefault}{\sfdefault}{m}{sl}
\SetMathAlphabet{\mathsfit}{bold}{\encodingdefault}{\sfdefault}{bx}{n}

\definecolor{cvprblue}{rgb}{0.21,0.49,0.74}
\usepackage[pagebackref,breaklinks,colorlinks,allcolors=cvprblue]{hyperref}
\hypersetup{
    colorlinks=true,
	citecolor=MidnightBlue,
	linkcolor=OrangeRed,
	urlcolor=OrangeRed
}
\usepackage{multirow}
\usepackage{makecell}
\usepackage{array}
\usepackage{amssymb}  
\usepackage{bbding}  
\usepackage{xcolor}
\usepackage[dvipsnames]{xcolor}
\usepackage{titletoc}
\usepackage{algorithm} 
\usepackage{algorithmic}
\usepackage{multicol}

\def\paperID{***} 
\def\confName{CVPR}
\def\confYear{2026}

\title{SFTok: Bridging the Performance Gap in Discrete Tokenizers}

\newcommand*\samethanks[1][\value{footnote}]{\footnotemark[#1]}
\author{Qihang Rao\thanks{Equal contribution.}, Borui Zhang\samethanks , Wenzhao Zheng, Jie Zhou, Jiwen Lu\thanks{Corresponding author.}\\
Department of Automation, Tsinghua University, China
}

\begin{document}
\maketitle
\begin{abstract}
Recent advances in multimodal models highlight the pivotal role of image tokenization in high-resolution image generation.
By compressing images into compact latent representations, tokenizers enable generative models to operate in lower-dimensional spaces, thereby improving computational efficiency and reducing complexity.
Discrete tokenizers naturally align with the autoregressive paradigm but still lag behind continuous ones, limiting their adoption in multimodal systems.
To address this, we propose \textbf{SFTok}, a discrete tokenizer that incorporates a multi-step iterative mechanism for precise reconstruction.
By integrating \textbf{self-forcing guided visual reconstruction} and \textbf{debias-and-fitting training strategy}, SFTok resolves the training-inference inconsistency in multi-step process, significantly enhancing image reconstruction quality.
At a high compression rate of only 64 tokens per image, SFTok achieves state-of-the-art reconstruction quality on ImageNet (rFID = 1.21) and demonstrates exceptional performance in class-to-image generation tasks (gFID = 2.29). \footnote{\url{https://github.com/Neur-IO/SFTok}}

\end{abstract}    
\vspace{-1mm}
\section{Introduction} \label{sec:intro}
In recent years, image generation models have achieved remarkable achievements, 
enabling the synthesis of highly realistic images from natural language descriptions and reference images. 
Leading models, including HunyuanImage 3.0~\citep{cao2025hunyuanimage}, Seedream 4.0~\citep{seedream2025}, Nano Banana~\citep{nanobanana}, GPT-Image~\citep{gptimage}, and Emu 3.5~\citep{cui2025emu3}, have demonstrated the capability to generate complex scenes and artistic images, attracting widespread attention from both academia and industry.
To unify image and text generation, researchers have proposed joint training paradigms such as Transfusion~\citep{zhou2024transfusion}, which was further scaled up in subsequent work~\citep{cao2025hunyuanimage}.
However, these hybrid frameworks still combine the diffusion model's $l_2$ loss with the cross-entropy loss for textual data, which limits their training simplicity and framework generalization.
Therefore, we contemplate: \emph{can we construct a fully native multimodal unified training paradigm that relies solely on cross-entropy loss?} 
This approach can also benefit from the mature advancements in the auto-regressive domain, such as KV Cache~\citep{ge2023model}, leading to significant improvements in inference efficiency. 
Hence, discrete tokenizers need to be re-emphasized as a critical component.

\begin{figure}[t]
    \centering
    \begin{subfigure}[t]{0.258\textwidth}
        \centering
        \includegraphics[width=\textwidth]{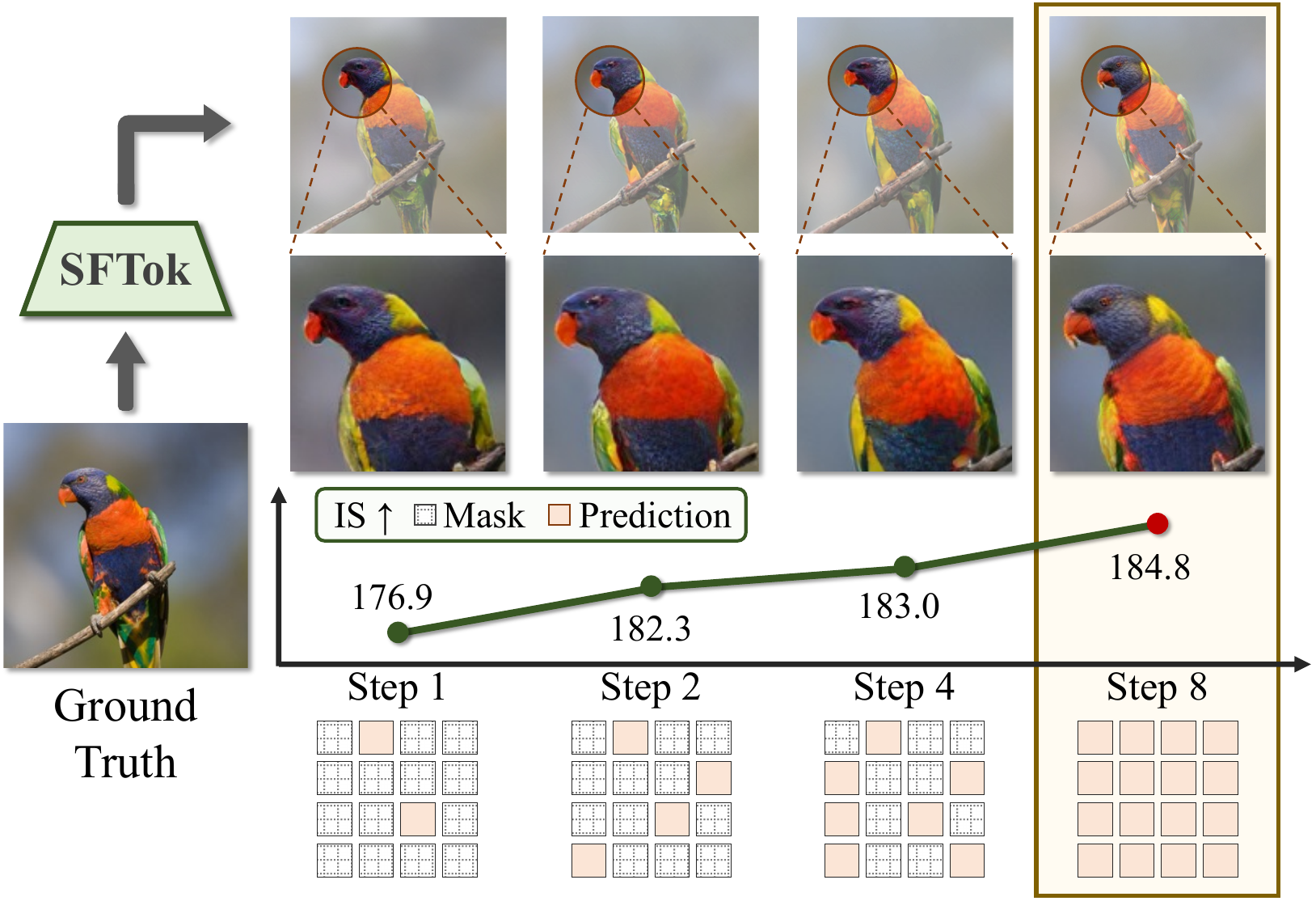}
        \caption{}
        \label{fig:head-a}
    \end{subfigure}
    \hfill
    \begin{subfigure}[t]{0.212\textwidth}
        \centering
        \includegraphics[width=\textwidth]{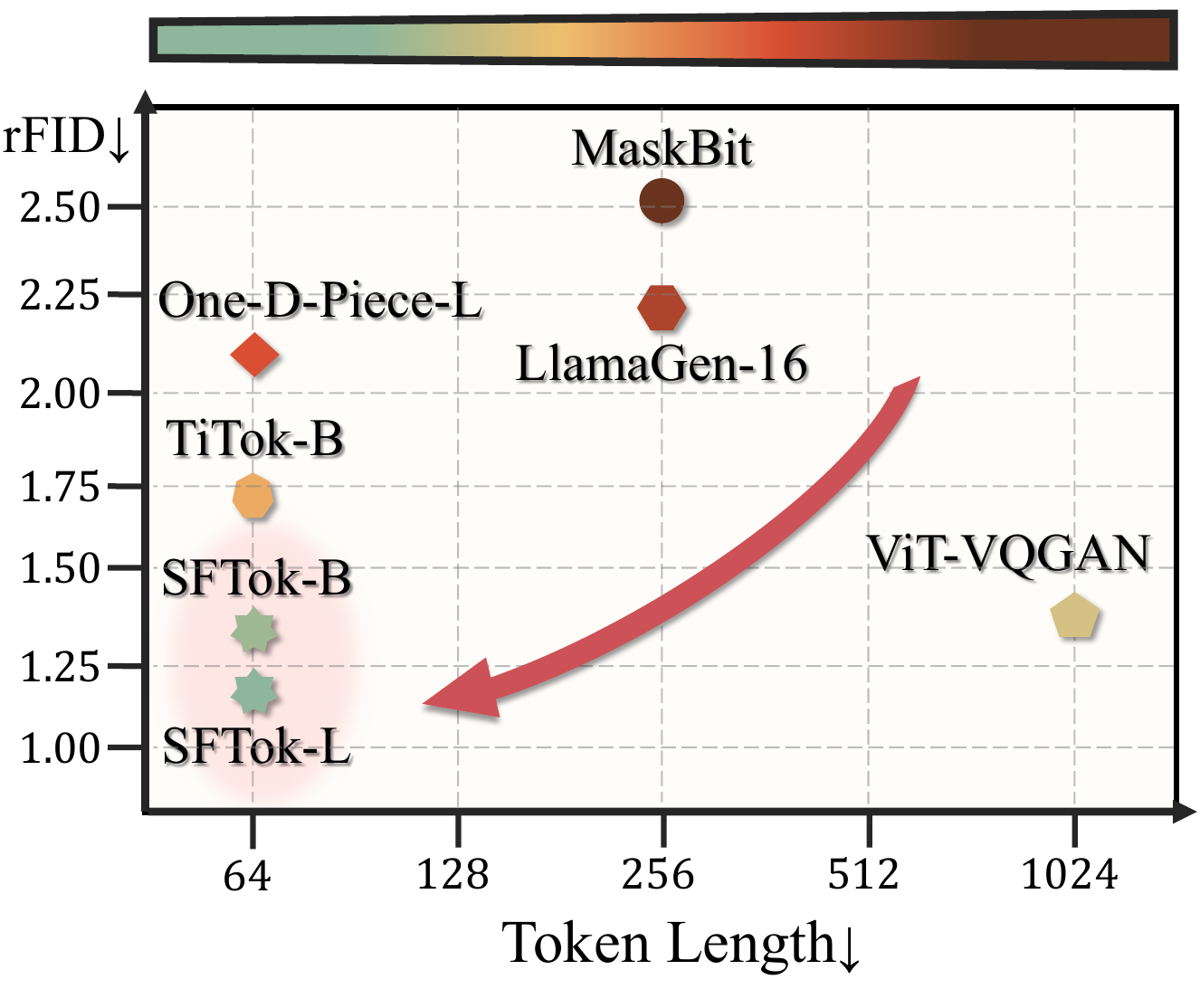}
        \caption{}
        \label{fig:head-b}
    \end{subfigure}
    \caption{SFTok employs self-forcing guided visual reconstruction (SFVR) that mitigates the training-inference inconsistency in multi-step iterative modeling within discrete tokenizers.
    (a) The image reconstruction quality progressively improves with the increase of inference steps.
    (b) High-quality image reconstruction is achieved at only 64 tokens (rFID = 1.21), outperforming other SOTA methods by a notable margin.}
    \label{fig:head}
    \vspace{-2mm}
\end{figure}

The image tokenizers employed in multimodal models can be broadly categorized into two types: 
\textbf{continuous tokenizers}~\citep{ho2020denoising, dhariwal2021diffusion, rombach2022high} and 
\textbf{discrete tokenizers}~\citep{esser2021taming, sun2024autoregressive}. 
Continuous tokenizers typically model Gaussian distributions, whereas discrete tokenizers, analogous to those used in text generation, model multinomial distributions, making them more naturally aligned with linguistic representations.
Due to training instability and higher compression rates, the reconstruction capability of discrete tokenizers is often inferior to that of continuous tokenizers, which greatly limits their application in multimodal model training. 
To enhance the performance of discrete tokenizers,
we draw inspiration from the \textbf{multi-step iterative} denoising process of diffusion models that utilize continuous tokenizers,
and seek to adapt this principle to discrete latent spaces.
Compared to traditional continuous tokenizers, diffusion models generate images by progressively denoising, gradually reducing the error in multiple iterations. 
These iterative steps can be regarded as decomposing the direct distribution prediction task into a sequence of conditional distribution prediction tasks.
Theoretically, such a formulation tends to achieve lower cross-entropy during prediction.
Previous works have also explored similar multi-step iterative mechanisms in discrete spaces, such as MaskGIT~\citep{chang2022maskgit}, which adopts a multi-step prediction method during the training of generative models.
In this process, the model iteratively predicts the tokens utilized for generation.
However, we observe that directly applying this multi-step iterative mechanism to discrete tokenizers fails to substantially enhance their reconstruction performance.
The underlying reason lies in the inconsistency between the training and inference processes, which introduces representational errors and hinders effective knowledge transfer from training to inference.
Therefore, ensuring consistency between training and inference in discrete space becomes a critical challenge.

\begin{figure}[t]
    \centering
    \begin{subfigure}[t]{0.261\textwidth}
        \centering
        \includegraphics[width=\textwidth]{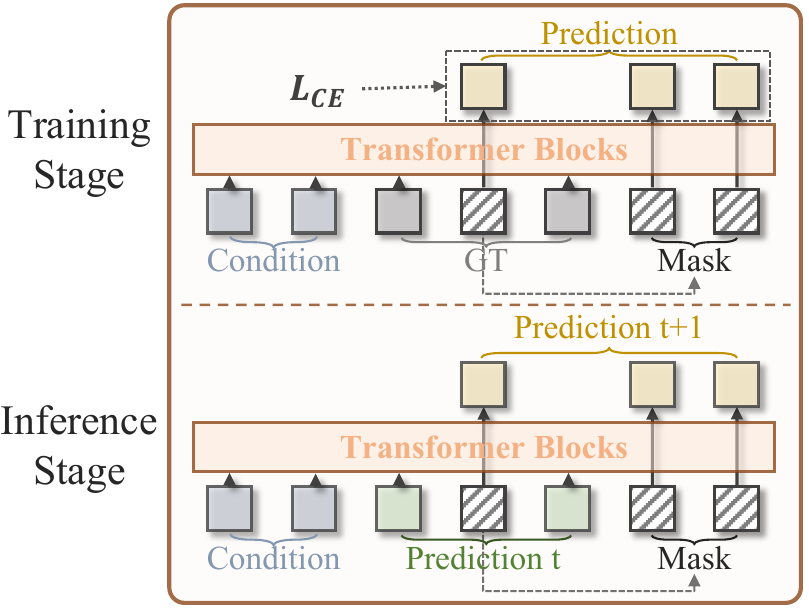}
        \caption{Vanilla strategy}
        \label{fig:replace_a}
    \end{subfigure}
    \hfill
    \begin{subfigure}[t]{0.208\textwidth}
        \centering
        \includegraphics[width=\textwidth]{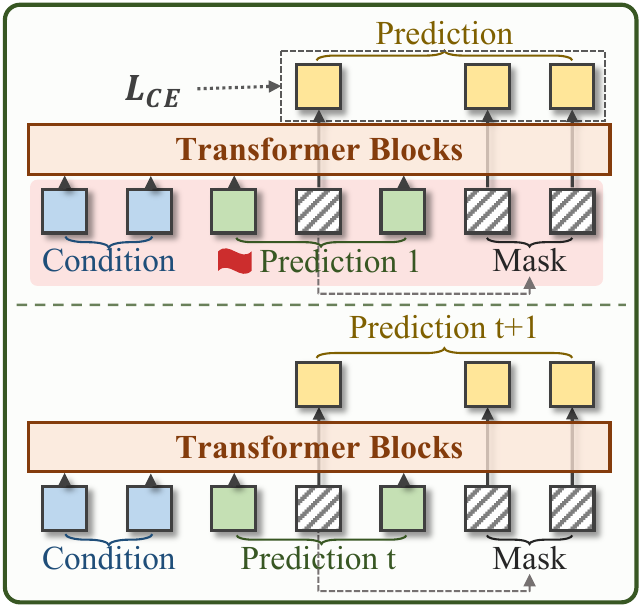}
        \caption{SFVR strategy}
        \label{fig:replace_b}
    \end{subfigure}

    \caption{Different mask token replacement strategies. (a) Vanilla strategy, where some masked tokens are replaced with the ground truth $\vm_g$. (b) Our proposed SFVR training strategy, where some masked tokens are replaced with the model's first-step prediction $\hat{\vm}_1$, ensuring training-inference consistency.}
    \label{fig:replace}
    \vspace{-1mm}
\end{figure}

In this paper, we propose \textbf{SFTok}, a novel discrete tokenizer framework that employs multi-step iterative modeling to enhance reconstruction quality, as demonstrated in \cref{fig:head}.
By comparing the predicted distributions with the ground truth at each step, 
we observe that although the model is trained to align predictions with the ground truth, a noticeable discrepancy persists after convergence.
Conventional training paradigms fail to account for this discrepancy, leading to an inconsistency between training and inference processes, as illustrated in \cref{fig:replace}.
To bridge this gap, we introduce a \textbf{self-forcing guided visual reconstruction (SFVR)} strategy, and formulate a dedicated \textbf{debias-and-fitting training} scheme.
At a high compression rate of 64 tokens per image, SFTok achieves state-of-the-art reconstruction quality on ImageNet (rFID = 1.21) and demonstrates excellent performance in downstream generative tasks (gFID = 2.29).
The core contributions are as follows:
\begin{itemize}
  \item \textbf{Identification of training-inference inconsistency}: We identify that conventional multi-step training strategies neglect the distribution discrepancy between training and inference, leading to a fundamental inconsistency.
  \item \textbf{Visual Condition Bias Correction}: We propose self-forcing guided visual reconstruction and debias-and-fitting training to resolves the inconsistency.
  \item \textbf{Discrete Tokenizer with Strong Capability}: We propose SFTok, which achieves superior performance in both image reconstruction and generation tasks.
\end{itemize}
\section{Related Work} \label{sec:relatedwork} 

Image tokenizers, by providing a computationally efficient and simpler alternative to high-dimensional pixel space, 
have become a foundational component in multimodal models for both image understanding and generation.
Current mainstream methods can be roughly divided into \textbf{continuous} and \textbf{discrete} categories.

\vspace{-2mm}
\paragraph{Continuous Tokenizers.}

Continuous tokenizers can be traced back to the Variational Autoencoder (VAE) and its variants~\citep{kingma2013auto,higgins2017beta}, 
which compress image representations by mapping the image space into a low-dimensional Gaussian latent space. 
Building upon this foundation, DC-VAE~\citep{parmar2021dual} introduces an instance-level discriminative loss along with a set-level adversarial loss, 
which enhances the representational capacity of the model without requiring any architectural modifications. 
DC-AE~\citep{chen2024deep} further improves the reconstruction quality under high spatial compression ratios 
through two key techniques: residual autoencoding and decoupled high-resolution adaptation. 
In contrast, DiTo~\citep{chen2025diffusion} integrates the diffusion framework into the design of continuous tokenizers, 
where the use of diffusion $l_2$ loss ensures high-quality image reconstruction and downstream generation performance. 
Due to their strong representational power in continuous latent spaces, 
continuous tokenizers have been widely adopted in diffusion-based image generation tasks~\citep{chen2023pixart,ma2024sit,podell2023sdxl}. 
By leveraging the gradual denoising process inherent to diffusion models, 
they are capable of generating high-quality images under diverse conditioning signals. 
However, the continuous nature of their representations makes them inherently incompatible 
with auto-regressive generation paradigms, 
thereby limiting their applicability in large-scale multimodal generative models. 
Consequently, increasing research attention has been directed toward the development of discrete tokenizers.

\vspace{-1mm}
\paragraph{Discrete Tokenizers.}
The original VQ-VAE~\citep{van2017neural} pioneered the use of discrete latent representations for images. 
However, early discrete tokenizers often suffered from limited reconstruction quality. 
To overcome this limitation, subsequent research made substantial progress by refining model architectures. 
VQ-VAE-2~\citep{razavi2019generating} introduced a multi-scale quantization strategy to preserve high-frequency details, 
while incorporating Vision Transformers~\citep{yu2022vector,cao2023efficient} further enhanced representational capacity.  
In parallel, improvements in objective functions also played a critical role in advancing reconstruction quality. 
VQGAN~\citep{esser2021taming} significantly enhanced the perceptual realism of reconstructed images by integrating GANs~\citep{goodfellow2014generative} and perceptual losses~\citep{larsen2016autoencoding,johnson2016perceptual}. 
Meanwhile, semantic supervision emerged as another effective paradigm. 
VAR~\citep{tian2024visual} leveraged DINO~\citep{oquab2023dinov2} as a semantic prior to improve reconstruction fidelity, 
while ImageFolder~\citep{liimagefolder} introduced a semantic branch within the quantization module, supervised by a contrastive loss. 
TiTok~\citep{yu2024image} employed a teacher–student discrete modeling strategy to improve reconstruction quality, 
while FlexTok~\citep{bachmann2025flextok} integrated continuous and discrete tokenizers to achieve high-fidelity and flexible image representations.
Nevertheless, the aforementioned discrete tokenizers are constrained by their one-step reconstruction mechanism, limiting both reconstruction performance and their potential in multimodal applications, which motivates the proposed \textbf{SFTok} framework. 
\vspace{-1mm}
\section{Method} \label{sec:method}
\vspace{-1mm}
In this section, we first present the key components and training framework of the 1D discrete tokenizer in \cref{sec:preliminary}.
We then analyze the training-inference discrepancy inherent in conventional multi-step iterative methods within discrete spaces, and propose the \textbf{self-forcing guided visual reconstruction} to mitigate in \cref{sec:method_sf}. 
Finally, we present the \textbf{debias-and-fitting training} protocol, designed to enhance the training stability of our proposed tokenizer in \cref{sec:three-stage}.

\subsection{Preliminaries: 1D Discrete Tokenizer Training} \label{sec:preliminary}
A compact latent representation is essential for handling high-dimensional data.
Autoencoders~\citep{hinton2006reducing} pioneered low-dimensional image encoding, while VAEs~\citep{kingma2013auto} advanced this concept by incorporating prior distributions, thereby enabling data generation through latent space sampling. 
With the emergence of the GPT era, discrete image representations have become essential to align with the discrete token-based frameworks of large language models. 
Discrete tokenizers~\citep{van2017neural}, which substitute continuous priors with discrete codebooks, have gained widespread adoption in image generation~\citep{esser2021taming,chang2022maskgit,rombach2022high} and large-scale pre-training~\citep{baobeit2022,bai2024sequential}.

\vspace{-2mm}
\paragraph{Mathematical Notation.}
Let $\mathcal{X} = \{\vx_i\}_{i=1}^N$ denote the image dataset. 
A standard 1D discrete tokenizer comprises an encoder $f_e(\cdot)$, a decoder $f_d(\cdot)$, and a quantizer $q(\cdot)$. 
Given an input image $\vx \in \mathbb{R}^{H \times W \times 3}$, 
a patch embedding layer first converts it into a sequence of image tokens $\vx_t \in \mathbb{R}^{L_1 \times D}$, 
where $L_1 = HW / P^2$ is the number of patches and $D$ is the embedding dimension. 
The encoder then applies transformers blocks jointly on the concatenated sequence of image tokens $\vx_t$ and query tokens $\vz \in \mathbb{R}^{K \times D}$. 
Through self-attention layers, query tokens extract and compress features from both the image tokens, yielding encoded query features $\mZ_e = f_e(\vx_t, \vz)$. 
For each vector $\vz_e \in \mathbb{R}^{D}$ in $\mZ_e$, the quantizer performs a nearest-neighbor search in the codebook $\mathcal{C} = \{\vc_1, \vc_2, \ldots, \vc_n\}$ to find the closest code vector:
\begin{equation}
    \vz_q = q(\vz_e) = \arg\min_{\vc_i \in \mathcal{C}} || \vz_e - \vc_i ||_2,
\end{equation}
yielding the quantized vector $\mZ_q = q(\mZ_e)$. 
The decoder then concatenates a mask token $\mM \in \mathbb{R}^{L_2 \times D}$ and $\mZ_q$ as inputs for decoding, producing $\hat{\mM} = f_d([\mM;\mZ_q])$ to extract the image features from $\mZ_q$.

\vspace{-2mm}
\paragraph{Discrete Space Cascaded Training.}
To achieve high compression ratios, cascaded training is employed in 1D tokenizers~\citep{yu2024image, miwa2025one}, where an additional teacher discrete tokenizer guides the training of the current discrete tokenizer.
Specifically, the input image $\vx$ is encoded and quantized by the frozen teacher model, which serve as the ground truth $\mM_g$ to supervise the current model.
At this stage, the training objective for $\hat{\mM}$ is to fit the ground truth $\mM_g$, i.e., to model the distribution $p(\mM_g | \mZ_q)$, rather than directly reconstructing the image.
The predicted tokens $\hat{\mM}$ are then decoded into an image by the teacher model's decoder, which functions as a pixel prediction head.
Typically, in later training stages, the teacher model's decoder is fine-tuned jointly with the target model to further enhance reconstruction quality.

\vspace{-2mm}
\paragraph{Single-Step Prediction.}
The 1D discrete tokenizer training paradigm adheres to a single-step prediction mechanism,
where the model directly predicts the ground truth $\mM_g$ from the quantized image tokens $\mZ_q$ by modeling $p(\mM_g | \mZ_q)$.
However, due to the substantial sequence length of mask tokens, direct distribution modeling becomes challenging.
Decomposing the single-step prediction into multiple steps can effectively alleviate this issue.
The autoregressive paradigm employs next-token prediction, transforming the single-step task into a series of conditional distribution predictions.
Similarly, the diffusion paradigm utilizes multi-step denoising to gradually refine noisy images into clear outputs.
The core challenge lies in adapting such multi-step iterative strategies for discrete image tokenizer training.
To address this challenge, we propose the SFTok framework.

\subsection{SFTok: Consistent Multi-Step Modeling} \label{sec:method_sf}

\begin{figure*}[ht]
    \centering
    \includegraphics[width=1\textwidth]{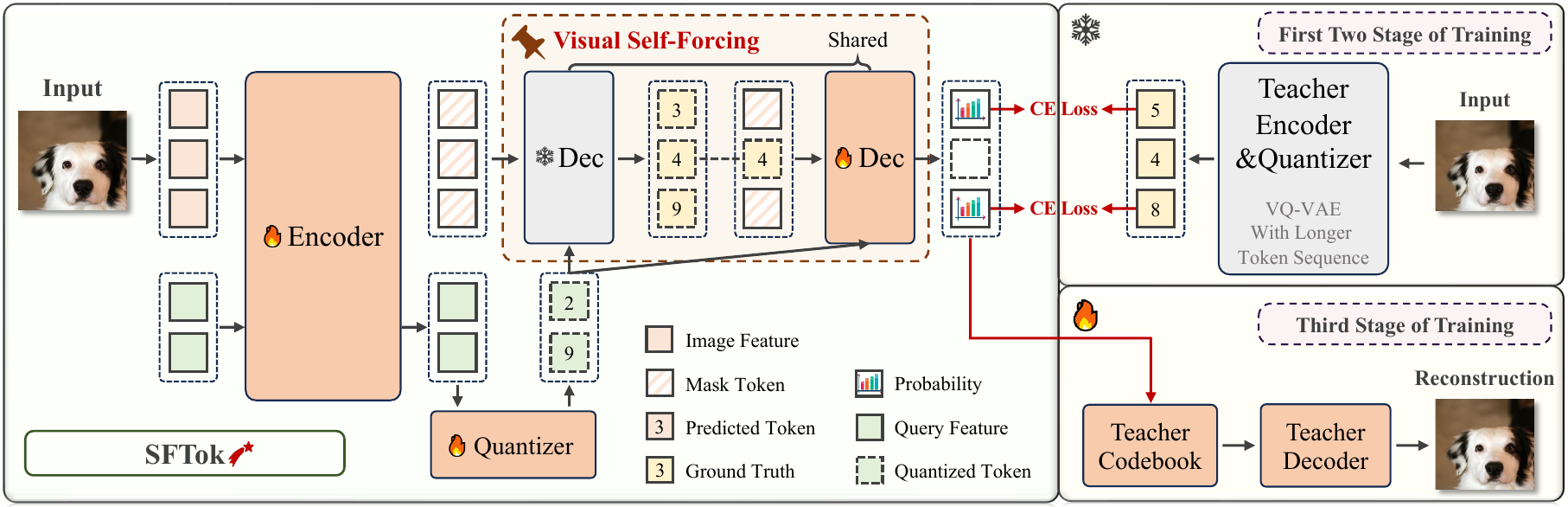}
    \caption{Model architecture for SFTok. SFTok consists of an encoder, a quantizer, a decoder and a teacher model.
    Image features, along with query features, are processed by the ViT encoder and quantized into discrete tokens.
    The decoder then predicts the masked tokens with discrete ones through a multi-step iterative process, and the teacher model converts the predicted tokens into reconstructed images.}
    \label{fig:framework}
    \vspace{-4mm}
\end{figure*}

\paragraph{Inconsistency between Training and Inference.}
We aim to adapt multi-step iterative approaches for discrete image tokenizer training
by modeling the conditional distribution:
\begin{align}
    p(\mM_{N} | \hat{\mM}_{\overline{N}}, \mZ_q) = p(\vm_{{i_1}}, \dots, \vm_{{i_N}} | \hat{\vm}_{j_1}, \dots, \hat{\vm}_{j_{\overline{N}}}, \mZ_q), \notag
\end{align}
where $\{i_1, \dots, i_N\} \cup \{j_1, \dots, j_{\overline{N}}\} = \{1, \dots, L_2\}$, 
$\mM_{N}$ denotes the set of $N$ mask token positions to be predicted, and $\hat{\mM}_{\overline{N}}$ represents the set of unmasked token positions that have already been predicted by the model.
In principle, multi-step conditional distribution modeling can reduce learning difficulty and improve image reconstruction quality.
Moreover, multi-step iterative methods typically achieve lower cross-entropy loss, implying smaller reconstruction errors, as demonstrated by the following derivation (see Appendix for detailed proof).
The cross-entropy is given by $ \text{CE}(p \| q) = \text{KL}(p \| q) + H(p) $.
For ideal single-step reconstruction, the KL divergence converges to zero, and the minimum loss corresponds to the entropy:
\begin{equation} \label{equ:single-step}
    L_{\min}^{s} = \sum_{i=1}^{L_2} H(\vm_i | \mZ_q).
\end{equation}
For multi-step conditional distribution prediction, assuming the KL divergence also approaches zero, we obtain:
\begin{equation} \label{equ:multi-step}
    L_{\min}^{m} = \sum_{i=1}^{L_2} H(\vm_i | \mZ_q, \vm_{\backslash i}^{pred}),
\end{equation}
where $\vm_{\backslash i}$ denotes the set of all tokens excluding $\vm_i$, 
and $\vm_{\backslash i}^{pred}$ represents the predicted values of $\vm_{\backslash i}$.
Defining the conditional mutual information as 
$I(\vm_i; \vm_{\backslash i}^{pred} | \mZ_q) = H(\vm_i | \mZ_q) - H(\vm_i | \mZ_q, \vm_{\backslash i}^{pred})$, we derive:
\begin{align} \label{equ:mutual_info}
    & H(\vm_i | \mZ_q, \vm_{\backslash i}^{pred}) = H(\vm_i | \mZ_q) - I(\vm_i; \vm_{\backslash i}^{pred} | \mZ_q) \notag \\
    & L_{\min}^{m} = \sum_{i=1}^{L_2} \left[ H(\vm_i | \mZ_q) - I(\vm_i; \vm_{\backslash i}^{pred} | \mZ_q) \right].
\end{align}
Since mutual information is non-negative, it follows that
$L_{\min}^{m} \leq L_{\min}^{s}$,
which indicates that multi-step distribution prediction can achieve a lower minimum loss compared to single-step prediction.

Implementing multi-step conditional distribution training and prediction in discrete space remains a key challenge.
We observe that existing discrete-space multi-step prediction methods, such as MaskGIT~\citep{chang2022maskgit},
while effective for generation tasks, yield suboptimal results when directly applied to image reconstruction.
Experimental analysis reveals the root cause: \textbf{inconsistency between training and inference processes},
leading to modeling errors and impeding knowledge transfer from training to inference.
Specifically, MaskGIT-style methods employ a random mask replacement during training,
where some token positions are randomly replaced with ground truth $\mM_g$
to simulate gradual masked token prediction during inference.
However, during inference, the model lacks access to the ground truth $\mM_g$ and must rely on its own predictions $\hat{\mM}_{\overline{i}}$, 
where $\hat{\vm}_{\overline{1}}, \dots, \hat{\vm}_{\overline{T}}$ are concatenated to form $\hat{\mM}$ for subsequent predictions, with $T$ denoting the total number of steps.
Crucially, the model's predictions $\hat{\mM}_{\overline{i}}$ often deviate from the ground truth $\mM_g$.
This discrepancy, unaccounted for during training, becomes progressively amplified during multi-step inference,
resulting in training-inference inconsistency and inferior performance compared to single-step prediction.

\vspace{-3mm}
\paragraph{Self-Forcing Guided Visual Reconstruction.}
We propose that fundamentally addressing training-inference inconsistency requires a mask replacement strategy that accurately simulates the distributional discrepancy between model predictions and ground truth during training.
As illustrated in the $T = 8$ example, at each step $i$, we first use the current model to predict the masked tokens, obtaining the predictions $\hat{\mM}_{\overline{i}}$.
We then compare the distribution discrepancies between the predictions at each step $\hat{\vm}_{\overline{1}}, \dots, \hat{\vm}_{\overline{T-1}}$, the ground truth $\mM_g$, and the final prediction $\hat{\mM}_{\overline{T}}$, along with the prediction's Top-1 accuracy.
The distribution discrepancy is measured by KL divergence.

Empirical results in \cref{fig:kl} reveal a substantial gap between the ground truth $\mM_g$ and the final prediction $\hat{\mM}_{\overline{T}}$, both in distribution and Top-1 accuracy. 
This gap is postulated as the fundamental cause of the training-inference inconsistency. 
Additionally, it is observed that with an increasing number of prediction steps, the distribution discrepancy between $\hat{\mM}_i$ and $\hat{\mM}_{\overline{T}}$ gradually reduces, concurrent with a rise in Top-1 accuracy. 
The distribution of $\hat{\mM}_{\overline{1}}$ is found to already closely approximate that of $\hat{\mM}_{\overline{T}}$.

Based on these observations, we propose a new mask replacement strategy, termed self-forcing guided visual reconstruction (SFVR), to mitigate the training-inference inconsistency.
Specifically, rather than replacing some masked tokens with the ground truth $\mM_g$, we first perform a forward pass without accumulating gradients, using the model to obtain the prediction $\hat{\mM}_{\overline{1}}$ at the first step.
Then, we replace some masked tokens with $\hat{\mM}_{\overline{1}}$, thereby better simulating the distributional characteristics of the model’s predictions during inference.
It is worth noting that, although the ideal replacement strategy would be to use $\hat{\mM}_{\overline{N-1}}$ to simulate the generation at the $N$-th step, 
we choose to use $\hat{\mM}_{\overline{1}}$ for replacement due to its distribution being already very close to that of the final prediction, offering a more computationally efficient solution.
By using the SFVR strategy, the input distributions during training and inference are better aligned, alleviating the training-inference inconsistency and improving the performance of multi-step prediction.

\subsection{Debias-and-Fitting Training of SFTok} \label{sec:three-stage}
This section outlines the debias-and-fitting training strategy of SFTok, which is composed of three stages: “warming up,” “distribution alignment modeling,” and “fine-tuning.”
In the “warming up” and “distribution alignment modeling” stages, the decoder of MaskGIT~\citep{chang2022maskgit} is used as a pixel prediction head, with its parameters frozen.
The “fine-tuning” stage involves jointly fine-tuning the decoder of MaskGIT to enhance the reconstruction quality.

\begin{figure}[t]
    \centering
    \begin{subfigure}[t]{0.235\textwidth}
        \centering
        \includegraphics[width=\textwidth]{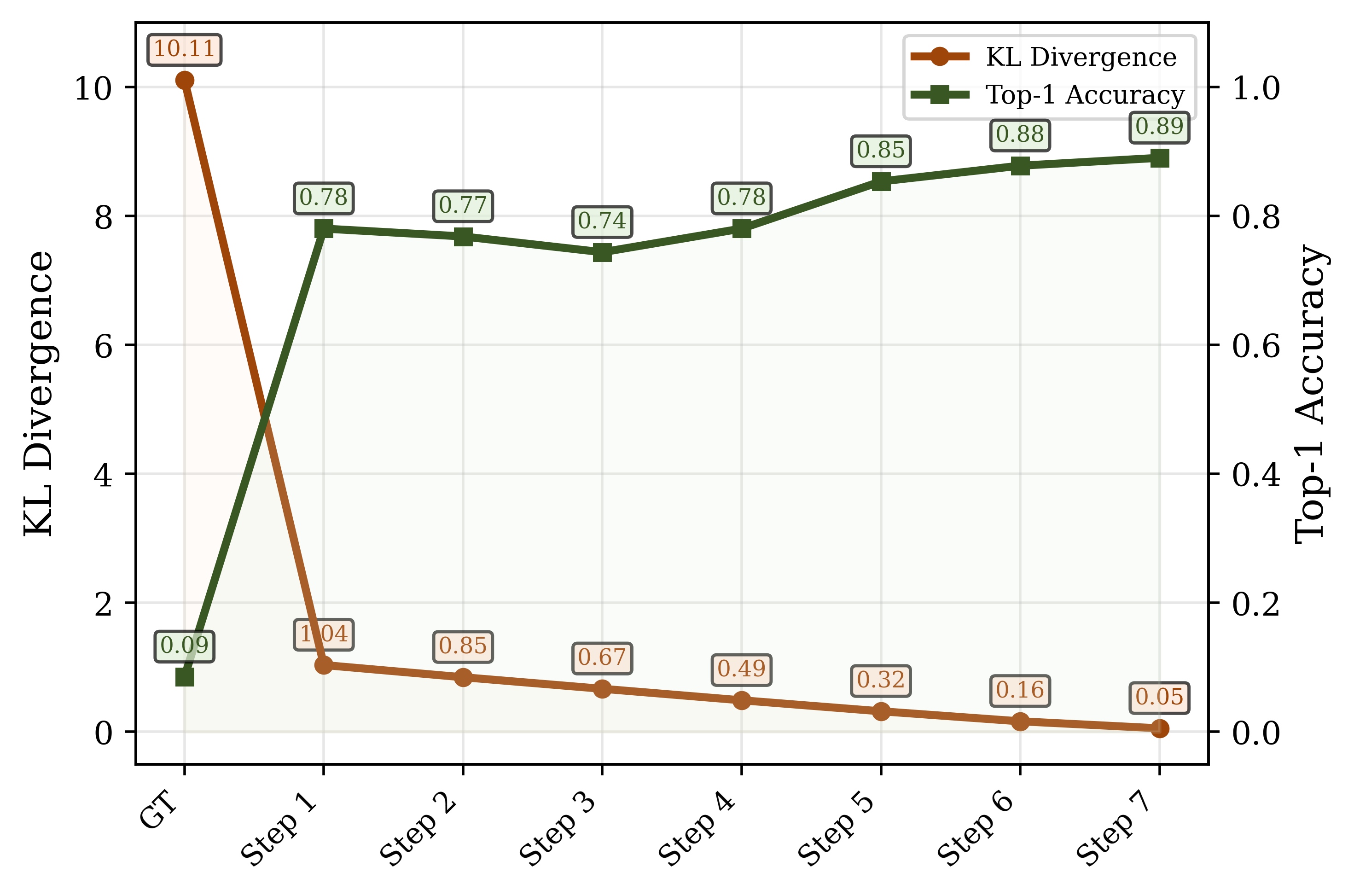}
        \caption{KL Div and Top-1 Acc.}
        \label{fig:kl}
    \end{subfigure}
    \hfill
    \begin{subfigure}[t]{0.235\textwidth}
        \centering
        \includegraphics[width=\textwidth]{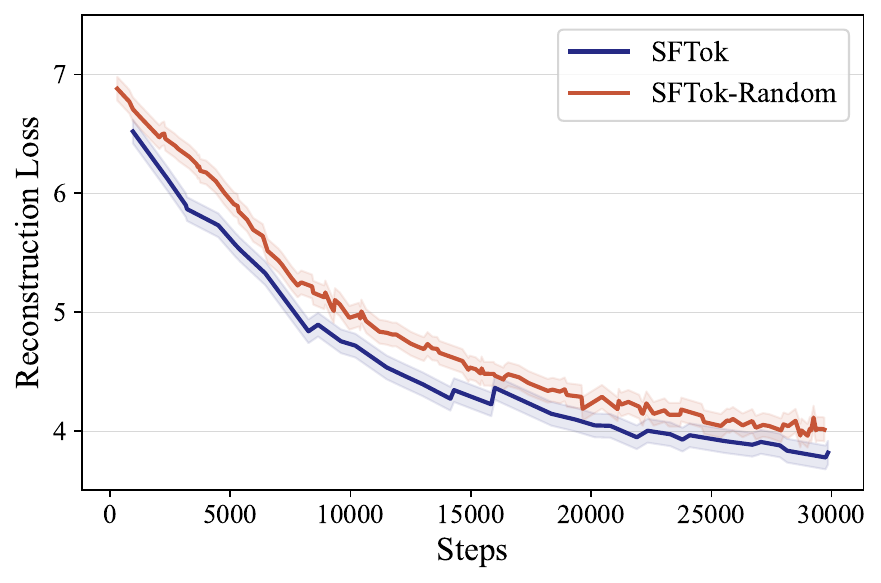}
        \caption{Reconstruction loss curves.}
        \label{fig:loss_curve}
    \end{subfigure}
    \caption{(a) Visualization of KL divergence and Top-1 accuracy at each prediction step \( \hat{\vm}_i \), compared to the ground truth \( \vm_g \) and the final prediction \( \hat{\vm}_T \) during multi-step prediction. This illustrates the core cause of the training-inference inconsistency.
        (b) Variation curves of reconstruction loss with and without warming-up training, demonstrating the necessity of warming-up.}
    \label{fig:kl_and_loss}
    \vspace{-2mm}
\end{figure}

\vspace{-3mm}
\paragraph{Training with Teacher Model.}
For both the "warming up" and "distribution alignment modeling" stages, the decoder from MaskGIT is adopted as the pixel prediction head.
Mirroring TiTok~\citep{yu2024image}, the pre-trained MaskGIT~\citep{chang2022maskgit} is used as a teacher model, from which the SFTok model learns a fitted distribution to bolster training stability and accelerate convergence.
The "warming up" stage specifically simulates single-step prediction by replacing no mask tokens, focusing on improving the initial prediction's accuracy.
As shown in \cref{fig:loss_curve}, which presents the variation curves of reconstruction loss with and without warm-up training: when warm-up training is not performed, the first-step prediction of the model lacks semantic information,amounting to random guessing. 
This increases the difficulty of model training,leading to significantly poorer convergence of metrics such as reconstruction loss,thus demonstrating the necessity of warm-up training.
In the “distribution alignment modeling” stage, we introduce the SFVR strategy proposed earlier, replacing some mask tokens with the model’s first-step prediction, thereby enabling multi-step conditional distribution training and prediction, which enhances image reconstruction quality.

\vspace{-3mm}
\paragraph{Fine-tuning Stage.}
After the first two training stages, we optionally proceed with the “fine-tuning” stage to improve the reconstruction quality.
In this stage, we freeze the encoder and quantizer of SFTok, and only train the decoder of both SFTok and pretrained MaskGIT towards pixel space using the typical VQGAN training approach.
We observe that this debias-and-fitting training strategy significantly improves training stability and reconstructed image quality, as demonstrated by the experimental results in~\cref{sec:comparison}.

\vspace{-3mm}
\paragraph{Image Generation with MaskTok.}
In addition to the reconstruction task, we also evaluate the performance of SFTok in image generation tasks.
Specifically, we adopt MaskGIT~\citep{chang2022maskgit} as our generation framework and replace its discrete tokenizer with SFTok to train a generative model.
The experimental setup is consistent with that in MaskGIT, which is briefly outlined below.
The image is pre-tokenized into 1D discrete tokens, and during training, a random proportion of the tokens is replaced by mask tokens. 
A bidirectional transformer is then applied to predict the masked tokens.
During the inference phase, multiple iterative steps are performed. 
In each step, the model predicts the masked tokens based on the unmasked tokens and updates a portion of the masked tokens with the predicted values. 
Finally, the predicted discrete tokens are decoded back into the pixel space using the well-trained SFTok decoder.
\section{Experiment} \label{sec:exp}

\begin{figure*}[t]
    \centering
    \includegraphics[width=0.96\textwidth]{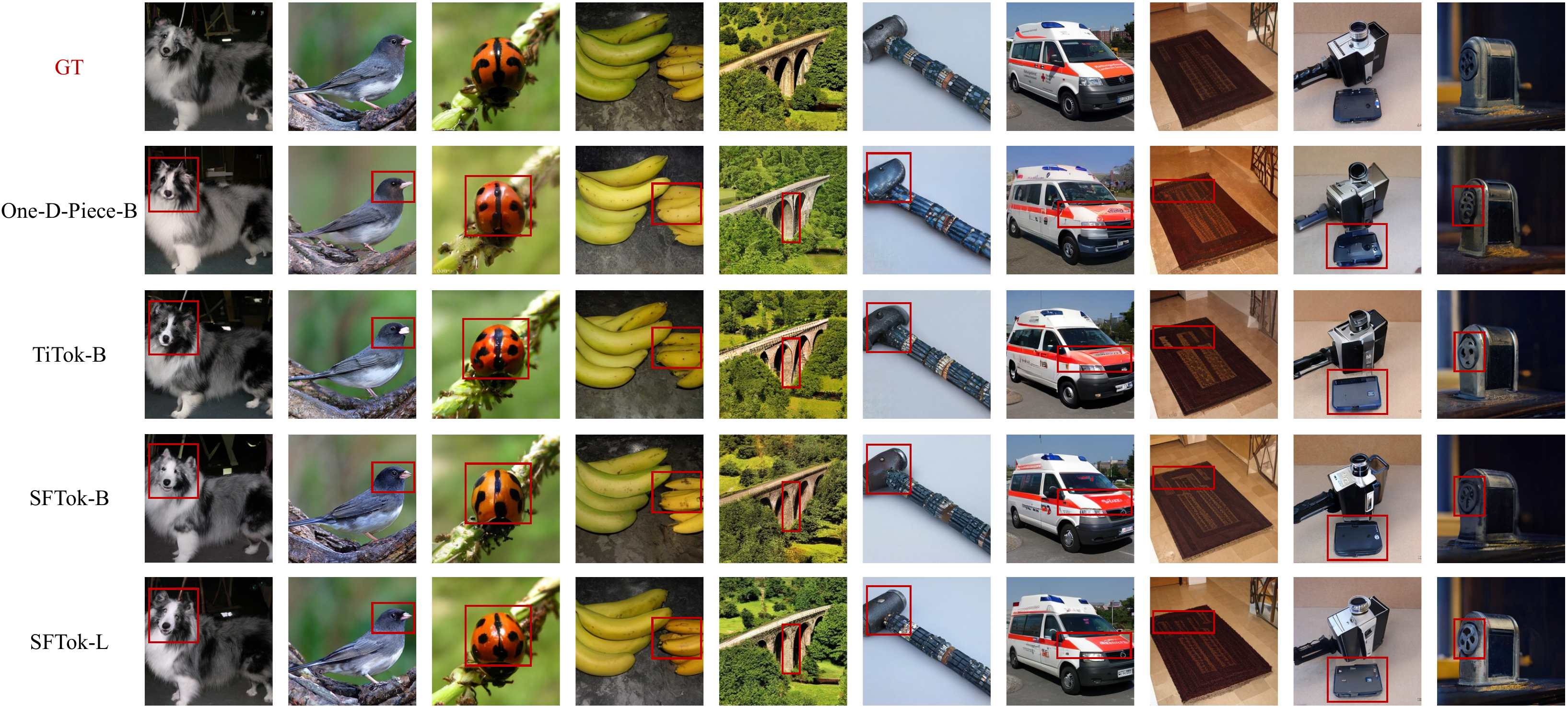}
    \caption{Visualizations of reconstruction results on the ImageNet validation set, with detailed comparisons highlighted in \textbf{\textcolor[HTML]{8B0000}{red}} boxes.}
    \label{fig:recon}
    \vspace{-1mm}
\end{figure*}

In this section, we present a comprehensive set of experiments to substantiate the effectiveness of SFTok framework.
We begin by comparing the performance of SFTok with state-of-the-art discrete tokenizers on the ImageNet dataset in \cref{sec:comparison}, followed by an evaluation of SFTok’s performance on downstream generative tasks in \cref{sec:generation}.
\cref{sec:ablation} presents an ablation study to analyze the contributions of different components within SFTok, validating the effectiveness of the self-forcing guided visual reconstruction strategy and the debias-and-fitting training protocol.

\subsection{Experimental Detail} \label{sec:exp_detail}

\paragraph{Model Setting.}
Our model is divided into two scales based on parameter size: SFTok-B and SFTok-L.
Both scales consist of an encoder, quantizer, decoder, and a pixel prediction head, encoding a $256 \times 256$ image from ImageNet~\citep{deng2009imagenet} into 64 tokens.
Similar to TiTok~\citep{yu2024image}, we use ViT~\citep{dosovitskiy2020image} as the underlying architecture for both the encoder and decoder.
The encoder and decoder of SFTok-B are both based on ViT-B, while SFTok-L uses ViT-B for the encoder and ViT-L for the decoder.
To ensure stable training and efficient codebook utilization, we employ optvq~\citep{zhang2024preventing} as the quantizer, with a codebook size of 8192 for both scales.
The pixel prediction head uses the decoder from MaskGIT~\citep{chang2022maskgit}.

\vspace{-1mm}
\paragraph{Optimizer Setting.}
The implementation of all models was carried out using the PyTorch~\citep{paszke2019pytorch} framework, with training conducted on a single machine equipped with eight NVIDIA 4090 GPUs.
AdamW~\citep{loshchilov2019decoupled} was used as the optimizer throughout the training process.
Due to GPU memory constraints, the batch size for training in stage 1 and stage 2 was set to 32 per GPU, while stage 3 employed a batch size of 8 per GPU.
The learning rate for all three stages was set to $1 \times 10^{-4}$, with a cosine decay learning rate scheduler applied.
Stage 1 training was conducted for 200k steps, stage 2 for 800k steps, and stage 3 for 1000k steps.

\subsection{Performance Comparison} \label{sec:comparison}

\begin{table}[htbp] \small
  \vspace{-1mm}
  \centering
  \caption{Quantitative comparison with state-of-the-art methods on the ImageNet dataset. †: Results presented on GitHub.}
  \label{tab:imagenet-recon}
  \begin{tabular}{|l|l|l|c|}
  \hline
  \textbf{Method} & \textbf{Token Length} & \textbf{\#Code} & \textbf{rFID↓}\\
  \hline
        ViT-VQGAN~\citep{yu2022vector} & 1024 \tiny{(16$\times$16)} & 8,192 & 1.28 \\
        Mo-VQGAN~\citep{zheng2022movq} & 1024 \tiny{(16$\times$16$\times$4)} & 1,024 & 1.12 \\
        ImageFolder~\citep{liimagefolder} & 572 \tiny{(286$\times$2)} & 4,096 & 0.80 \\
  \hline
        VQGAN~\citep{esser2021taming} & 256 \tiny{(16$\times$16)} & 16,384 & 3.64 \\
        MaskGIT~\citep{chang2022maskgit} & 256 \tiny{(16$\times$16)} & 1,024 & 2.28 \\
        RQ-VAE~\citep{lee2022autoregressive} & 256 \tiny{(8$\times$8$\times$4)} & 16,384 & 3.20  \\
        MaskBit~\citep{weber2024maskbit} & 256 \tiny{(16$\times$16)} & 4,096 & 1.61  \\
        COSMOS~\citep{agarwal2025cosmos} & 256 \tiny{(16$\times$16)} & 64,000 & 2.52  \\
        VQGAN-LC~\citep{zhu2024scaling} & 256 \tiny{(16$\times$16)} & 100,000 & 2.62  \\
        LlamaGen-L~\citep{sun2024autoregressive} & 256 \tiny{(16$\times$16)} & 16,384 & 2.19 \\
  \hline
        TiTok-B~\citep{yu2024image}  & 64 \tiny{(1d-tokenizer)}   & 4,096 & 1.70  \\
        TiTok-BL\textsuperscript{†}~\citep{yu2024image} & 64 \tiny{(1d-tokenizer)}   & 8,192 & 2.06  \\
        One-D-Piece-B~\citep{miwa2025one} & 64 \tiny{(1d-tokenizer)}   & 4,096 & 2.39  \\
        One-D-Piece-L~\citep{miwa2025one} & 64 \tiny{(1d-tokenizer)}   & 4,096 & 2.10  \\
        \textbf{SFTok-B}   & \textbf{64} \tiny{(1d-tokenizer)}   & \textbf{8,192} & \textbf{1.44}  \\
        \textbf{SFTok-L}   & \textbf{64} \tiny{(1d-tokenizer)}   & \textbf{8,192} & \textbf{1.21}  \\
  \hline
  \end{tabular}
\end{table}

      
We perform a comparative evaluation of our SFTok model against state-of-the-art discrete tokenizers on the ImageNet dataset~\citep{deng2009imagenet}. 
Evaluations on the validation set employs rFID~\citep{heusel2017gans}, with results summarized in \cref{tab:imagenet-recon}. 
As shown, SFTok-B and SFTok-L achieve rFID scores of 1.44 and 1.21, respectively, in 8-step reconstruction when compressing images to only 64 tokens.
These results establish a new state-of-the-art in reconstruction quality at the present compression rate. 
Notably, SFTok-L outperforms many models with relatively lower compression rates, underscoring the efficiency of our approach.
For visualization, we compare several high-performance methods for compressing images to 64 tokens. 
Here’s a revised version of that sentence to avoid ambiguity and merge the two ideas smoothly:
Among them, the One-D-Piece-B method~\citep{miwa2025one}, which is compatible with variable token counts, is also evaluated using 64 tokens for reconstruction to ensure a fair comparison.
The results are shown in \cref{fig:recon}, further demonstrating the superiority of SFTok.
As highlighted in the red boxes, SFTok preserves fine-grained details in complex textures more effectively than alternative methods.

\subsection{Generation Evaluation} \label{sec:generation}


We evaluate the image generation performance of SFTok using the MaskGIT paradigm~\citep{chang2022maskgit}, with the model parameter configuration consistent with the original MaskGIT. 
However, in contrast to the MaskGIT method, we utilize the SFTok-B and SFTok-L models, which are trained via the debias-and-fitting strategy, as discrete tokenizers.
The generative model generates different token sequences based on different class labels, and the sequences are subsequently converted into images through the SFTok decoder.
In the generation process, we adopt the same sampling strategy as MaskGIT and utilize 8-step iterative generation for image reconstruction. The model’s class-specific image generation performance is then assessed on ImageNet~\citep{deng2009imagenet}.
As shown in \cref{tab:generation}, both SFTok-B and SFTok-L-based generation models outperform MaskGIT and all other transformer-based generation models listed in the table in terms of gFID.
Additionally, SFTok demonstrates superior performance compared to diffusion-based generation models.
Furthermore, the qualitative results of the generated images, as presented above, highlight SFTok’s remarkable ability to preserve fine-grained details and textures.

\begin{table}[t]
    \centering
    \caption{Generation Comparsion on ImageNet.}
    \label{tab:generation}
    \begin{tabular}{|l|c|c|}
    \hline
    \textbf{Method} & \textbf{\#Type} & \textbf{gFID↓}\\
    \hline
    LDM-4~\citep{rombach2022high} & diffusion  & 3.60 \\
    DC-AE~\citep{chen2024deep}    & diffusion  & 1.88  \\
    \hline
    Open-Magvit2-B~\citep{yulanguage}         & transformer  & 3.08 \\
    LlamaGen-L~\citep{sun2024autoregressive}  & transformer  & 3.80 \\
    MaskGIT~\citep{chang2022maskgit}          & transformer  & 6.18 \\
    TiTok-B-64-MaskGIT~\citep{yu2024image}    & transformer  & 2.48 \\
    \hline
    SFTok-B-MaskGIT   & transformer  & 2.32 \\
    SFTok-L-MaskGIT   & transformer  & 2.29 \\
    \hline
    \end{tabular}
    \vspace{-1mm}
\end{table}

\subsection{Ablation Study} \label{sec:ablation}

In this section, we conduct an ablation study to analyze the contributions of different components within SFTok, thoroughly validating the effectiveness of the SFVR and the debias-and-fitting training protocol. 
We also investigate the impact of different mask replacement ratios on model performance, demonstrating the importance of fully simulating the inference distribution during the training phase to enhance model performance. 
Additionally, we evaluate the model’s performance across different reconstruction step counts to highlight SFTok’s superiority in multi-step iterative reconstruction and justify our choice of 8-step reconstruction as the optimal setting.

\vspace{-1mm}
\paragraph{Training-Inference Consistency Training.}

We begin our analysis by investigating the impact of the proposed self-forcing guided visual reconstruction (SFVR) strategy and evaluating the effectiveness of the warming-up procedure.
In addition to SFTok-B, we train two separate ablation models on ImageNet: one using the baseline replacement strategy and the other employing our proposed approach. 
For both models, all configurations are maintained identical, except for the inclusion or exclusion of the warming-up phase.
To ensure a fair comparison, each model is trained for 1000k steps, which corresponds to the total number of steps in the first two stages of SFTok-B.
As demonstrated in \cref{tab:ablation_dar}, the SFVR strategy results in a significant improvement in reconstruction quality compared to the vanilla method.
Moreover, the inclusion of the warming-up procedure further enhances the overall performance of the model, providing additional evidence of its importance for training.

\begin{figure}[t]
    \centering
    \includegraphics[width=0.47\textwidth]{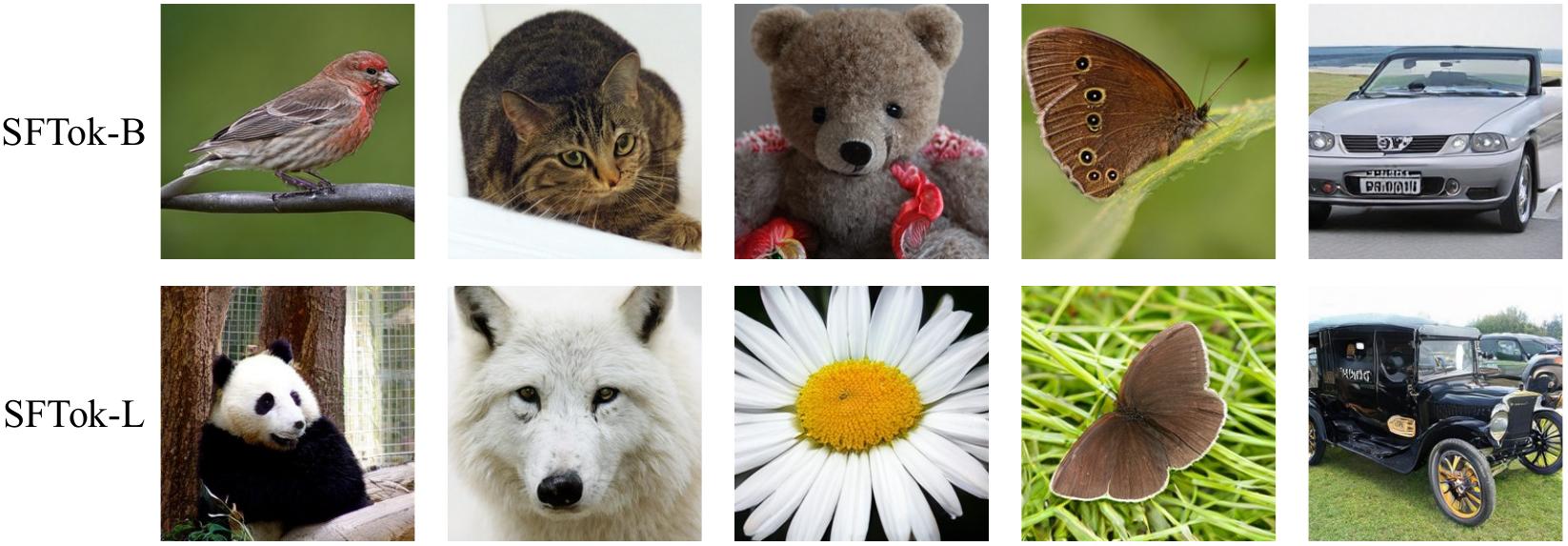}
    \caption{Visualizations of generation results.}
    \label{fig:generate}
\end{figure}

\begin{table}[t]
      \small
      \centering
      \caption{Ablation study on SFVR and warming-up strategy.}
      \label{tab:ablation_dar}
      \begin{tabular}{|l|c|c|c|}
      \hline
      \textbf{Training strategy} & \textbf{\#Token} & \textbf{Warming-up} & \textbf{rFID↓} \\
      \hline
      Vanilla & 64 & \scriptsize{\XSolidBrush}    & 6.47 \\
      \hline
      SFVR & 64 & \scriptsize{\XSolidBrush}  & 4.40  \\
      SFVR & 64 & \scriptsize{\CheckmarkBold} & 4.33  \\
      \hline
      \end{tabular}
      \vspace{-1mm}
\end{table}

\vspace{-1mm}
\paragraph{Mask Replacement Ratio.}

We further investigate the effect of different mask replacement ratios during training on the model’s performance.
To this end, we conduct experiments with three distinct mask replacement ratios: 0.5, 0.8, and 1.0, while keeping all other training settings consistent with those of the SFTok-B configuration.
Specifically, we train stage 1 for 200k steps, and stage 2 for 800k steps.
As shown in \cref{tab:ablation_ratio}, the results indicate that a replacement ratio of 1.0 yields the best reconstruction quality.
This suggests that fully simulating the distribution encountered during inference during the training phase significantly enhances model performance.
By adopting this ratio, the model is able to better align its predictions with the inference distribution, effectively reducing the gap between training and inference behavior.
The improved performance highlights the importance of a more accurate simulation of inference conditions during training, which benefits the model’s ability to produce higher-quality reconstructions.

\paragraph{Image Reconstruction with Varying Steps.}

\begin{figure}[t]
    \centering
    \includegraphics[width=0.474\textwidth]{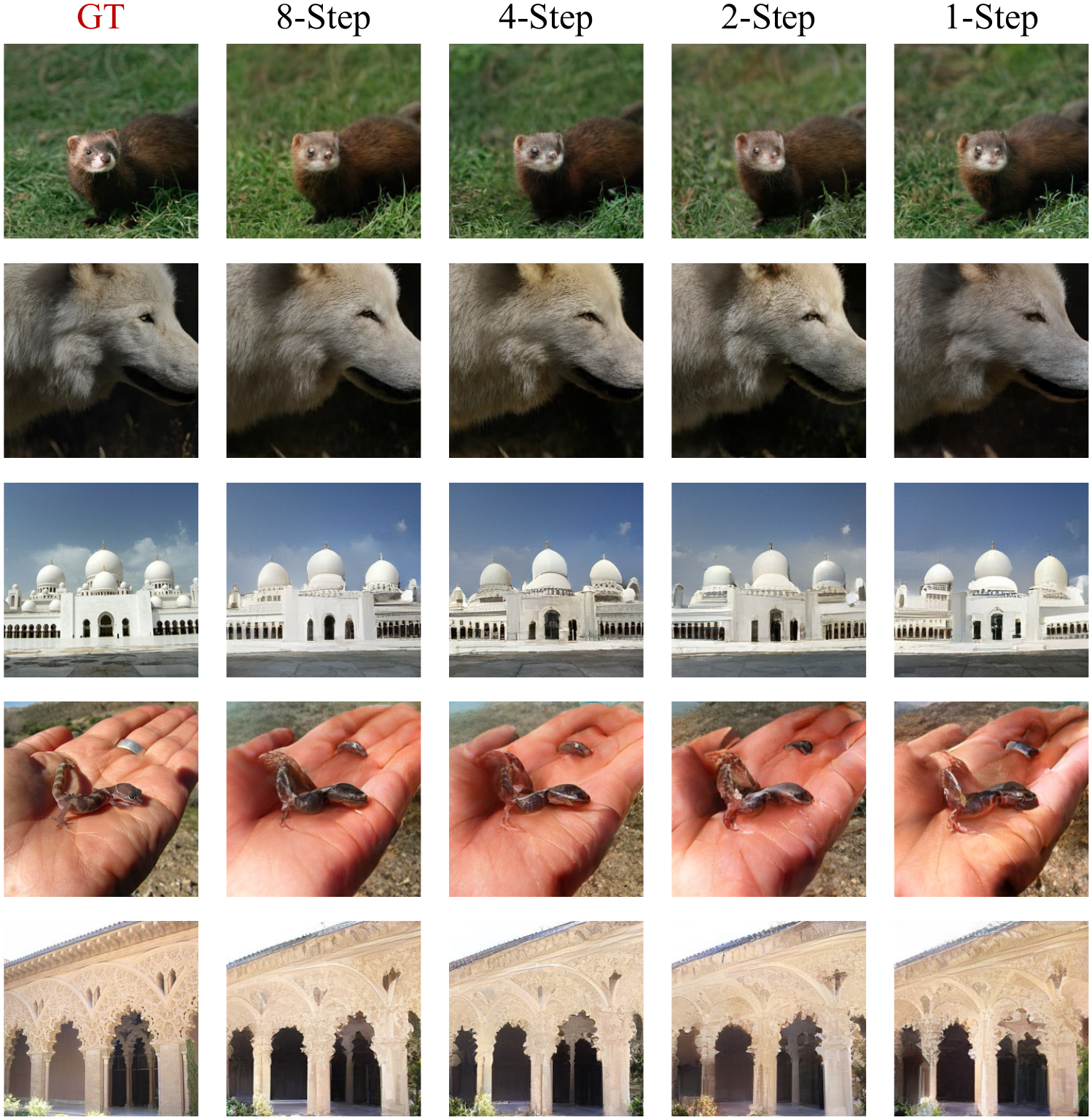}
    \caption{Visualizations of reconstruction results at different steps.}
    \label{fig:vis_diff_steps}
\end{figure}

\begin{figure}[t]
    \centering
    \includegraphics[width=0.475\textwidth]{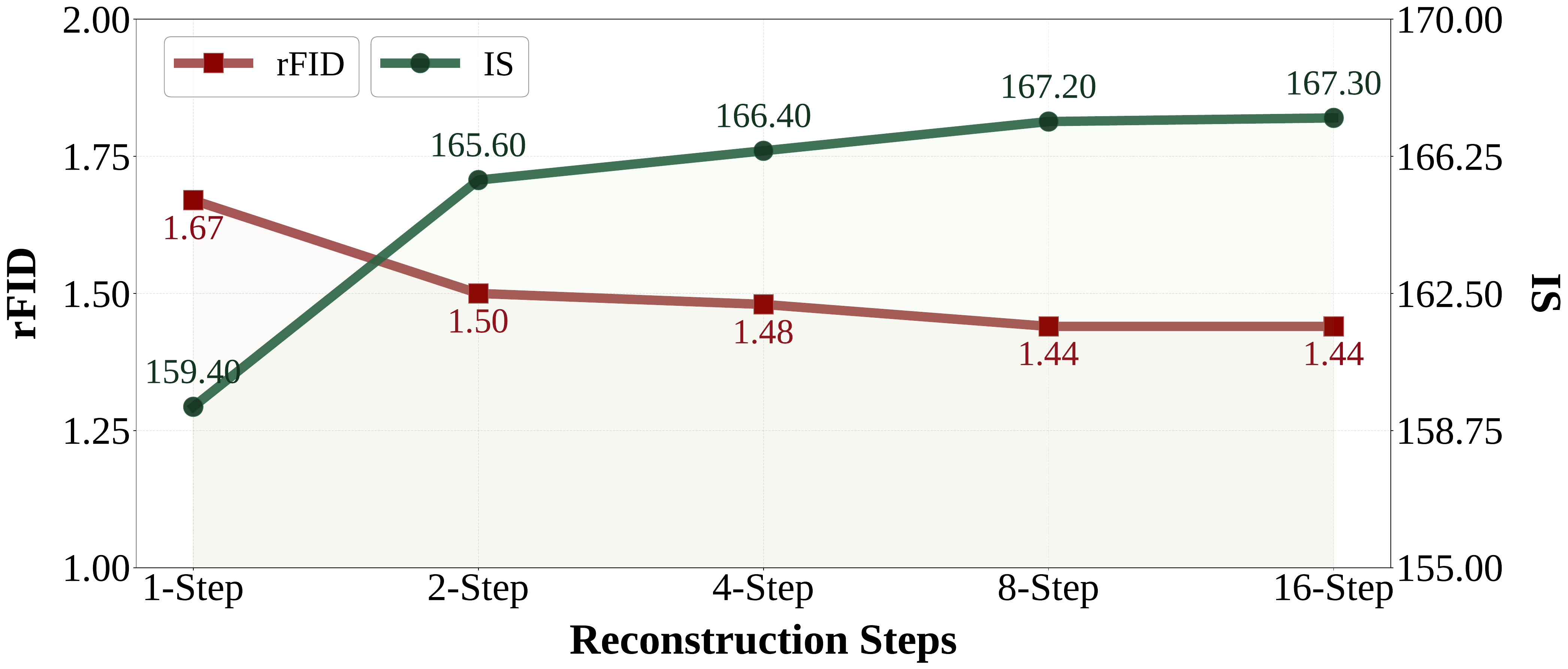}
    \caption{Quantitative reconstruction results at different steps.}
    \label{fig:diff_steps_rfid}
\end{figure}

%
We evaluate the image reconstruction performance of SFTok across different numbers of inference steps using a multi-step prediction strategy.
In this approach, the model iteratively predicts the masked tokens conditioned on the unmasked ones, progressively updating the tokens with the predicted values.
This strategy enables a systematic improvement in reconstruction quality, allowing us to compare performance across varying numbers of inference steps.
As shown in \cref{fig:diff_steps_rfid}, we conduct an ablation experiment using the SFTok-B model, which has undergone the debias-and-fitting training process, and evaluate its performance based on two quantitative metrics: rFID and IS.
As the number of inference steps increases from 1 to 8, the reconstruction quality improves significantly, demonstrating the effectiveness of the multi-step prediction strategy.
Furthermore, we present visualizations of the reconstruction results for different step numbers in \cref{fig:vis_diff_steps}.
To more clearly highlight the differences in detail across varying inference steps, we use the SFTok-B model trained through the first two stages for comparison.
As shown in \cref{fig:vis_diff_steps}, with an increasing number of inference steps, the details in the image become more refined, and the perceptual quality is enhanced.
Since the difference in reconstruction quality between the 16-step and 8-step results is marginal, we adopt 8-step as the default setting, striking a balanced trade-off between performance and efficiency.


\begin{table}[t]
      \small
      \centering
      \caption{Ablation study on mask replacement ratio.}
      \label{tab:ablation_ratio}
      \begin{tabular}{|l|c|c|c|}
      \hline
      \textbf{Model} & \textbf{Replacement Ratio} & \textbf{\#Token} & \textbf{rFID↓} \\
      \hline
      SFTok-B & 0.5   & 64 & 4.96 \\
      SFTok-B & 0.8   & 64 & 4.59  \\
      SFTok-B & 1.0   & 64 & 4.33  \\
      \hline
      \end{tabular}
\end{table}

\vspace{2mm}
\section{Discussion} \label{sec:conclusion}
\vspace{-1mm}

This paper focuses on addressing the performance gap in discrete tokenizers. 
Unlike single-step prediction, we draw inspiration from the multi-step iterative approach used in diffusion models for continuous tokenizers and aim to adapt it for the discrete space. 
However, we discovered that traditional training strategies fail to account for the distributional differences between training and inference, leading to the issue of training-inference inconsistency. 
This inconsistency hinders the effective application of multi-step iterations in the discrete space.
To resolve this challenge, we propose SFTok, which introduces a self-forcing guided visual reconstruction (SFVR) strategy to mitigate the training-inference inconsistency. 
Furthermore, we have designed a targeted, debias-and-fitting training strategy, which is conducted in three distinct stages to further enhance the effectiveness of this approach.
Through extensive experiments on the ImageNet dataset, we demonstrate the superior performance of SFTok in both image reconstruction and downstream generation tasks.
With 64 tokens, we achieved a high-quality reconstruction with an rFID of 1.21, and excellent performance in generation tasks (gFID = 2.29).

We also acknowledge that SFTok has some limitations. 
Our experiments were conducted at a resolution of $256 \times 256$, which demonstrated the efficiency of the SFTok training paradigm when compressed to 64 tokens. 
Due to computational constraints, we did not train or evaluate the model at higher resolutions or with larger models, nor did we explore the performance at higher compression rates.
In future work, we aim to extend the application of SFTok, exploring its performance at higher resolutions with larger models, as well as its applicability to more downstream tasks.
\newpage
{
    \small
    \bibliographystyle{ieeenat_fullname}
    \bibliography{main}
}

\onecolumn
\appendix
\clearpage

\onecolumn
\startcontents[sections]
\printcontents[sections]{l}{1}{
    \setcounter{tocdepth}{2}
    \section*{\centering Table of Content for Appendix \rule{\linewidth}{0.5pt}}
}
\rule{\linewidth}{0.5pt}

\section{Usage of Large Models in Paper Writing} \label{sec:llm_usage}
In the preparation of this manuscript, large language models (LLMs) were employed in two specific capacities: 

\textbf{Preliminary Literature Review}: During the initial literature survey, LLMs facilitated rapid familiarization with foundational concepts and recent developments in the field.
This automated exploration was subsequently complemented by systematic manual searches on arXiv and Google Scholar to verify and consolidate the obtained information.

\textbf{Writing Assistance}: Throughout the composition process, LLMs were utilized for grammatical verification, sentence refinement, and generation of selected descriptive passages.
All AI-generated content underwent rigorous review and editing to ensure compliance with academic standards and factual accuracy.

While recognizing the utility of LLMs in academic contexts, we also acknowledge their inherent limitations.
No LLMs were deployed in critical research components including code implementation, experimental design, or results interpretation.
These elements were exclusively developed by the authors.

The specific LLM implementation used was OpenAI's ChatGPT (GPT-4).
All usage complied with applicable ethical guidelines and platform policies to maintain academic integrity and original scholarship. 

\section{Computational Resources} \label{sec:compute_usage}
All experiments were conducted using 8 NVIDIA RTX 4090 GPUs with 24GB memory each for both training and evaluation. 

As shown in \cref{tab:compute_usage}, for the combined 1000k iterations of Stage 1 and Stage 2, the SFTok-B model required approximately 1224 GPU hours, while the SFTok-L model consumed approximately 1728 GPU hours. 
During Stage 3's 1000k iterations, the training time was approximately 1656 GPU hours for SFTok-B and 3264 GPU hours for SFTok-L.

\begin{table}[ht]
    \vspace{2mm}
    \centering
    \begin{tabular}{c|c|c|c}
        \hline
        \textbf{Model} & \textbf{Stage 1 + 2 (GPU hours)} & \textbf{Stage 3 (GPU hours)} & \textbf{Total (GPU hours)} \\
        \hline
        SFTok-B & 1224 & 1656 & 2880 \\
        SFTok-L & 1728 & 3264 & 4992 \\
        \hline
    \end{tabular}
    \caption{Computational requirements for model training.}
    \label{tab:compute_usage}
\end{table}

\vspace{2mm}
\section{Theoretical Analysis of Multi-Step Prediction} \label{sec:appendix_theory}
This section presents a theoretical analysis comparing multi-round masked conditional prediction against single-round direct prediction from both cross-entropy and accuracy perspectives.
Our analysis demonstrates that multi-round masking theoretically achieves a lower optimal expected loss and a higher theoretical accuracy upper bound compared to direct prediction.
The detailed derivations are provided below.

$Z_q$: Quantized vector; $M = \{m_1, m_2, ..., m_{L_2}\}$: Mask token, where $L_2$ is the number of tokens; 
$M_g$: Ground-truth of mask token;
$m_{\backslash i} = \{m_1, ..., m_{i-1}, m_{i+1}, ..., m_N\}$: Sequence excluding $m_i$; 
$m_{\backslash i}^{pred}$: Predicted subset of $m_{\backslash i}$; $p_{\theta}(m | Z_q)$: 
Model distribution ($\theta$: parameters); $\text{CE}(p \| q) = -\mathbb{E}_p[\log q]$, $\text{KL}(p \| q) = \mathbb{E}_p\left[\log \frac{p}{q}\right]$, $H(p) = -\mathbb{E}_p[\log p]$, $I(X; Y | Z) = H(X | Z) - H(X | Y, Z)$.

\vspace{1mm}
\subsection{Cross-Entropy Perspective}
\vspace{1mm}
\noindent \textbf{1. Relationship Between CE, KL Divergence, and Entropy}
\[
\text{CE}(p \| q) = \mathbb{E}_p\left[\log \frac{p}{q}\right] - \mathbb{E}_p[\log p] = \text{KL}(p \| q) + H(p)
\]
\[
\text{Since } \text{KL}(p \| q) \geq 0, \quad \text{CE}(p \| q) \geq H(p)
\]

\vspace{1mm}
\noindent \textbf{2. Direct Distribution Prediction (Single Iteration)}
\[
p_{\theta}(M | Z_q) = \prod_{i=1}^{L_2} p_{\theta, i}(m_i | Z_q) \quad (\text{Conditional independence assumption of tokens})
\]
\[
L_{\text{min}}^{s} = \text{CE}(M|Z_q) \| (M_g |Z_q) = H(M | Z_q) \quad(\text{KL}(p | p_{\theta}) = 0, \text{when the model distribution matches the true distribution})
\]
\[
H(M | Z_q) = \sum_{i=1}^{L_2} H(m_i | Z_q) \quad (\text{Factorization of joint entropy})
\]
\[
\Rightarrow L_{\text{min}}^{s} = \sum_{i=1}^{L_2} H(m_i | Z_q)
\]

\vspace{1mm}
\noindent \textbf{3. SFTok (Multiple Iteration)}
\[
p_{\theta}(M | Z_q) = \prod_{i=1}^{L_2} p_{\theta, i}(m_i | Z_q, m_{\backslash i}^{pred}) \quad (\text{Dependent on predicted tokens } m_{\backslash i}^{pred})
\]
\[
L_{\text{min}}^{m} = \sum_{i=1}^{L_2} H(m_i | Z_q, m_{\backslash i}^{pred}) \quad (\text{When } \text{KL}(p \| p_{\theta}) = 0)
\]
\[
\text{From } I(m_i; m_{\backslash i}^{pred} | Z_q) = H(m_i | Z_q) - H(m_i | Z_q, m_{\backslash i}^{pred})
\]
\[
\Rightarrow H(m_i | Z_q, m_{\backslash i}^{pred}) = H(m_i | Z_q) - I(m_i; m_{\backslash i}^{pred} | Z_q)
\]
\[
\Rightarrow L_{\text{min}}^{m} = \sum_{i=1}^{L_2} \left[ H(m_i | Z_q) - I(m_i; m_{\backslash i}^{pred} | Z_q) \right]
\]

\vspace{1mm}
\noindent \textbf{4. Loss Comparison Conclusion}
\[
L_{\text{min}}^{m} = L_{\text{min}}^{s} - \sum_{i=1}^{L_2} I(m_i; m_{\backslash i}^{pred} | Z_q)
\]
\[
\text{Since } I(m_i; m_{\backslash i}^{pred} | Z_q) \geq 0, \quad L_{\text{min}}^{m} \leq L_{\text{min}}^{s}
\]

\subsection{Accuracy Perspective}
\vspace{1mm}
\noindent \textbf{1. Definition of Bayes Optimal Top-1 Accuracy}
\[
\text{Acc}^*(m_i | Z_q) = \mathbb{E}_{p(m_i|Z_q)} \left[ \max_{x \in \mathcal{M}} p(m_i = x | Z_q) \right]
\]

\vspace{1mm}
\noindent \textbf{2. Derivation via Law of Total Probability + Jensen's Inequality}
\[
p(m_i = x | Z_q) = \mathbb{E}_{p(m_{\backslash i}^{pred} | Z_q)} \left[ p(m_i = x | Z_q, m_{\backslash i}^{pred}) \right]
\]
\[
\max_x p(m_i = x | Z_q) = \max_x \mathbb{E}\left[ p(m_i = x | Z_q, m_{\backslash i}^{pred}) | Z_q \right] \leq \mathbb{E}\left[ \max_x p(m_i = x | Z_q, m_{\backslash i}^{pred}) | Z_q \right]
\]

\vspace{1mm}
\noindent \textbf{3. Monotonicity Conclusion of Accuracy}
\[
\mathbb{E}_{p(M|Z_q)} \left[ \max_x p(m_i = x | Z_q) \right] \leq \mathbb{E}_{p(M|Z_q)} \left[ \mathbb{E}\left[ \max_x p(m_i = x | Z_q, m_{\backslash i}^{pred}) | Z_q \right] \right]
\]
\[
\Rightarrow \text{Acc}^*(m_i | Z_q) \leq \text{Acc}^*(m_i | Z_q, m_{\backslash i}^{pred})
\]
\vspace{1mm}
\noindent \textbf{4. Model Selection Implication}
\[
\text{Accuracy (Single-Iteration Model)} \leq \text{Accuracy (Multi-Iteration Model)}
\]
\vspace{1mm}

\section{Implementation Details} \label{sec:appendix_implementation}

\vspace{1mm}
\subsection{Datasets} \label{sec:appendix_datasets}

This study primarily utilizes the ImageNet dataset~\citep{deng2009imagenet}.
The training set consists of 1,281,167 images, while the validation set contains 50,000 images, both categorized into 1,000 classes.

\vspace{1mm}
\subsection{Configurations} \label{sec:appendix_config}
As a concrete example, we detail the configuration for the SFTok-B model. 
The SFTok-L model differs solely in its decoder, which employs a ViT-L architecture. 
The specific model and optimizer settings are shown in \cref{tab:appendix_config}.

\section{Additional Experiments} \label{sec:appendix_experiments}

\vspace{2mm}
\subsection{Detailed Results during Training} \label{sec:appendix_training}
\begin{figure}[ht]
    \centering
    \begin{minipage}[b]{0.45\linewidth}
        \centering
        \includegraphics[width=\linewidth]{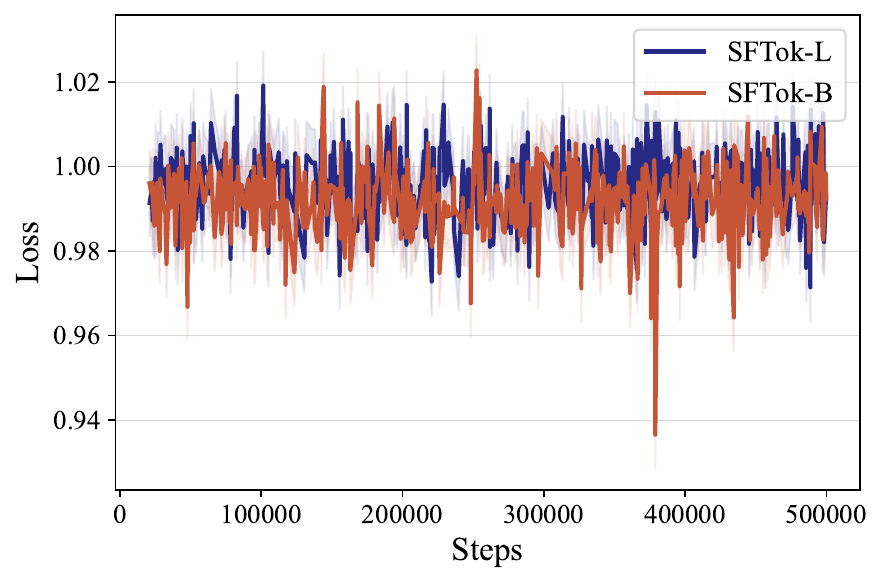}
        \caption{Discriminator loss curves for different models.}
        \label{fig:training_details_a}
    \end{minipage}
    \hfill
    \begin{minipage}[b]{0.45\linewidth}
        \centering
        \includegraphics[width=\linewidth]{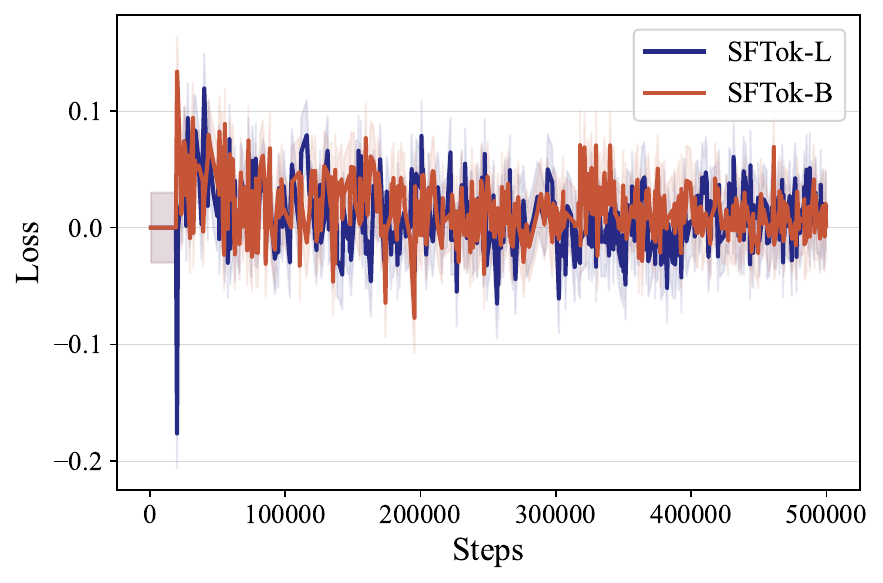}
        \caption{Gan loss curves for different models.}
        \label{fig:training_details_b}
    \end{minipage}
    
    \vspace{2mm}  

    \begin{minipage}[b]{0.45\linewidth}
        \centering
        \includegraphics[width=\linewidth]{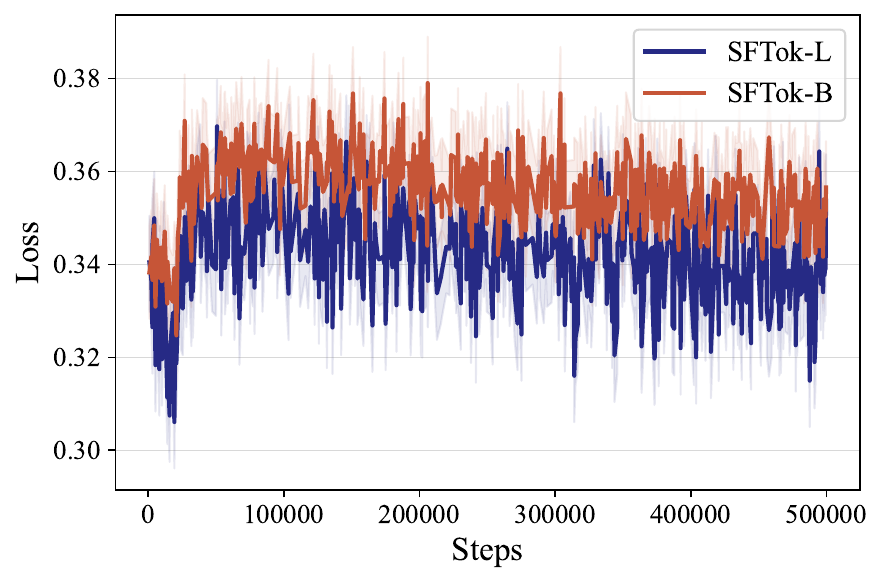}
        \caption{Perceptual loss curves for different models.}
        \label{fig:training_details_c}
    \end{minipage}
    \hfill
    \begin{minipage}[b]{0.45\linewidth}
        \centering
        \includegraphics[width=\linewidth]{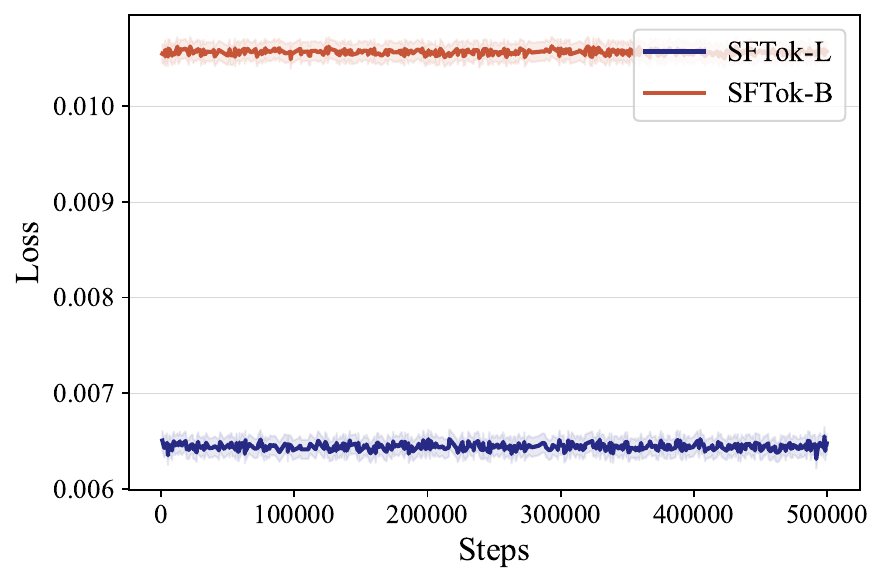}
        \caption{Quantizer loss curves for different models.}
        \label{fig:training_details_d}
    \end{minipage}

    \vspace{2mm}  

    \begin{minipage}[b]{0.45\linewidth}
        \centering
        \includegraphics[width=\linewidth]{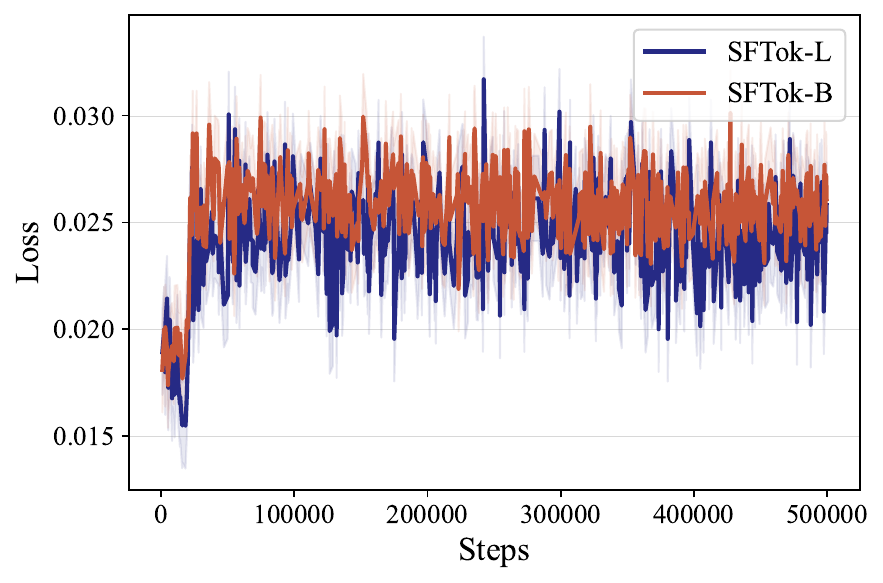}
        \caption{Reconstruction loss curves for different models.}
        \label{fig:training_details_e}
    \end{minipage}
    \hfill
    \begin{minipage}[b]{0.45\linewidth}
        \centering
        \includegraphics[width=\linewidth]{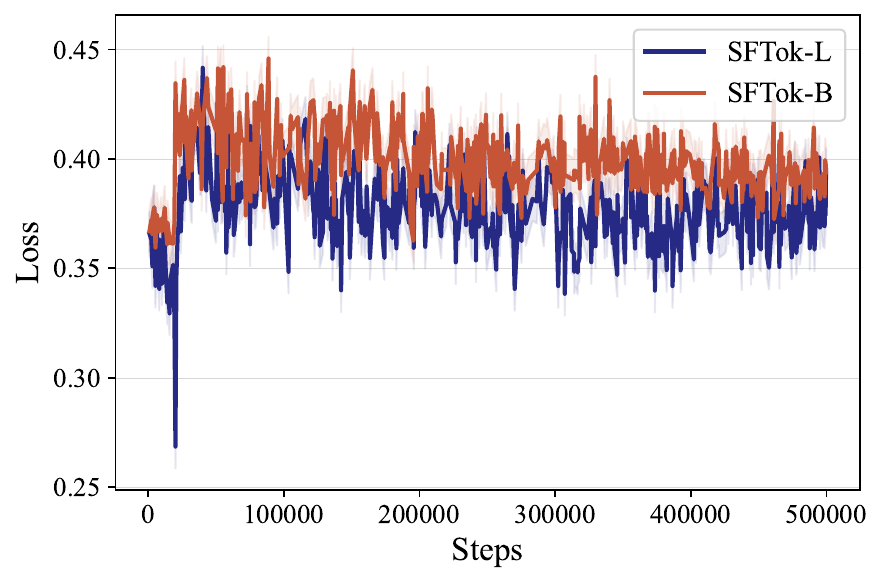}
        \caption{Total loss curves for different models.}
        \label{fig:training_details_f}
    \end{minipage}  
\end{figure}

We provide detailed loss curves for the first 500k iterations of the stage 3 training process for our SFTok-B and SFTok-L models in \cref{fig:training_details_a,fig:training_details_b,fig:training_details_c,fig:training_details_d,fig:training_details_e,fig:training_details_f}.
As shown in these figures, both SFTok-B and SFTok-L exhibit consistent decreasing trends across all loss metrics throughout training, indicating progressive optimization of their reconstruction capabilities.
Specifically, the abrupt change observed at 20k iterations is attributed to the introduction of discriminator and GAN losses. 
The quantizer loss remained stable because we froze the parameters of the encoder and quantizer during stage 3 training.
Furthermore, SFTok-L demonstrates superior performance over SFTok-B in all loss metrics during training, a finding that is subsequently validated through both quantitative and qualitative experiments.

\begin{table}[htp]
\centering
\small
\begin{tabular}{l|l|l|c|c|c}
\hline
\textbf{Cat.} & \textbf{Subcat.} & \textbf{Parameter} & \textbf{Stage 1} & \textbf{Stage 2} & \textbf{Stage 3} \\
\hline
\multirow{2}{*}{\textbf{Experiment}} & \multirow{2}{*}{-} & Max Train Examples & 1,281,167 & 1,281,167 & 1,281,167 \\
 & & Resume Training & True & True & True \\
\hline

\multirow{19}{*}{\textbf{Model}} & \multirow{9}{*}{\textbf{VQ Model}} & Codebook Size & 8,192 & 8,192 & 8,192 \\
 & & Token Size & 32 & 32 & 32 \\
 & & Use L2 Norm & True & True & True \\
 & & Commitment Cost & 0.25 & 0.25 & 0.25 \\
 & & Vit Enc Model Size & base & base & base \\
 & & Vit Enc Patch Size & 16 & 16 & 16 \\
 & & Num Latent Tokens & 64 & 64 & 64 \\
 & & Num Group & 1 & 1 & 1 \\
 & & Finetune Decoder & \textcolor{red}{False} & \textcolor{red}{False} & \textcolor{red}{True} \\
\cline{2-6}
 & \multirow{10}{*}{\textbf{Decoder}} & Vit Dec Model Size & base & base & base \\
 & & Vit Dec Patch Size & 16 & 16 & 16 \\
 & & Num Latent Tokens & 64 & 64 & 64 \\
 & & Token Size & 32 & 32 & 32 \\
 & & Num Proxy Codes & 256 & 256 & 256 \\
 & & Codebook Size & 1024 & 1024 & 1024 \\
 & & Randomize Temperature & 1.0 & 1.0 & 1.0 \\
 & & Guidance Scale & 0.0 & 0.0 & 0.0 \\
 & & Guidance Decay & "constant" & "constant" & "constant" \\
 & & Replace Prob & 1.0 & 1.0 & 1.0 \\
\hline

\multirow{12}{*}{\textbf{Losses}} & \multirow{12}{*}{-} & Quantizer Weight & 1.0 & 1.0 & 1.0 \\
 & & Label Smoothing & 0.0 & 0.0 & - \\
 & & Loss Weight Unmasked Token & 0.1 & 0.1 & - \\
 & & Discriminator Type & - & - & dinodisc \\
 & & Discriminator Start & - & - & 20,000 \\
 & & Discriminator Factor & - & - & 1.0 \\
 & & Discriminator Weight & - & - & 0.5 \\
 & & Perceptual Loss & - & - & lpips \\
 & & Perceptual Weight & - & - & 1.0 \\
 & & Reconstruction Loss & - & - & l2 \\
 & & Reconstruction Weight & - & - & 1.0 \\
 & & Lecam Regularization Weight & - & - & 0.001 \\
\hline

\multirow{4}{*}{\textbf{Dataset}} & \multirow{2}{*}{\textbf{Params}} & Num Workers Per GPU & 12 & 12 & 12 \\
 & & Dataset Type & simple & simple & simple \\
\cline{2-6}
 & \multirow{2}{*}{\textbf{Preprocessing}} & Resize Shorter Edge & 256 & 256 & 256 \\
 & & Crop Size & 256 & 256 & 256 \\
\hline

\multirow{5}{*}{\textbf{Optimizer}} & \multirow{5}{*}{\textbf{Params}} & Learning Rate & $1\times10^{-4}$ & $1\times10^{-4}$ & $1\times10^{-4}$ \\
 & & Discriminator Learning Rate & - & - & \textcolor{red}{$1\times10^{-4}$} \\
 & & Beta1 & 0.9 & 0.9 & 0.9 \\
 & & Beta2 & 0.99 & 0.99 & 0.999 \\
 & & Weight Decay & 1e-4 & 1e-4 & 1e-4 \\
\hline

\multirow{3}{*}{\textbf{Lr Scheduler}} & \multirow{3}{*}{\textbf{Params}} & Learning Rate & $1\times10^{-4}$ & $1\times10^{-4}$ & $1\times10^{-4}$ \\
 & & Warmup Steps & 10,000 & 10,000 & 10,000 \\
 & & End Lr & $1\times10^{-5}$ & $1\times10^{-5}$ & $1\times10^{-5}$ \\
\hline

\multirow{10}{*}{\textbf{Training}} & \multirow{10}{*}{-} & Gradient Accumulation Steps & 1 & 1 & \textcolor{red}{2} \\
 & & Per GPU Batch Size & 32 & 32 & \textcolor{red}{8} \\
 & & Mixed Precision & fp16 & fp16 & fp16 \\
 & & Enable TF32 & True & True & True \\
 & & Use EMA & True & True & True \\
 & & Max Train Steps & \textcolor{red}{200,000} & \textcolor{red}{800,000} & \textcolor{red}{1,000,000} \\
 & & Max Grad Norm & 1.0 & 1.0 & 1.0 \\
 & & Use Mlmloss & True & True & True \\
 & & Single Step Generation & \textcolor{red}{True} & \textcolor{red}{False} & False \\
 & & Guided Mask & \textcolor{red}{False} & \textcolor{red}{True} & True \\
\hline
\end{tabular}
\caption{Detailed parameter-level configuration comparison across three training stages}
\label{tab:appendix_config}
\end{table}

\vspace{2mm}
\subsection{More Reconstruction Results on ImageNet} \label{sec:appendix_imagenet}
\begin{figure}[htbp]
    \centering
    \vspace{-3mm}
    \includegraphics[width=0.65\linewidth]{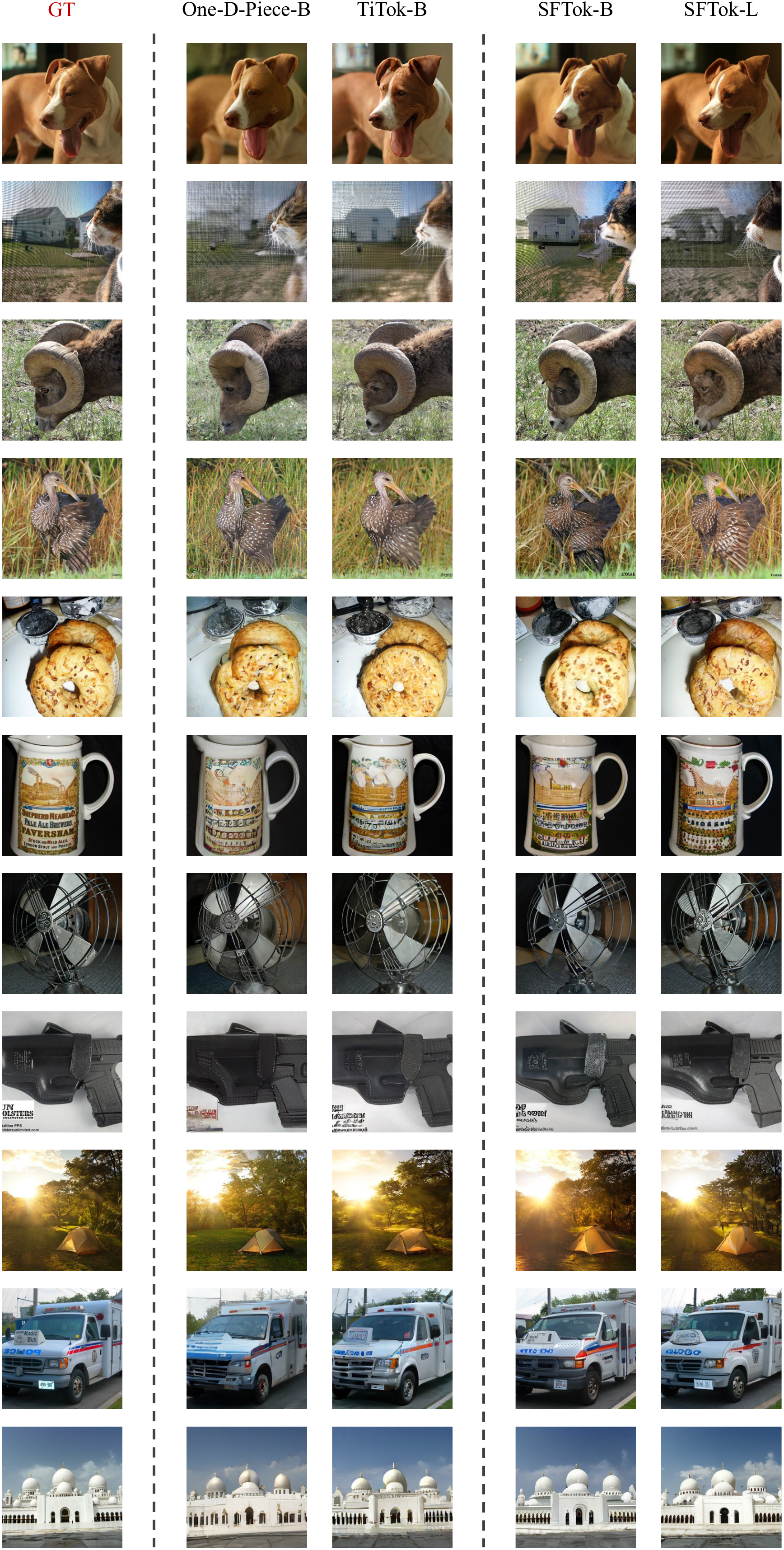}
    \caption{Additional reconstruction results on the ImageNet dataset.}
    \label{fig:appendix_recon}
\end{figure}

We provide additional reconstruction results in \Cref{fig:appendix_recon}, offering further visual evidence of our SFTok model's consistent superiority on ImageNet. 
The samples demonstrate its enhanced capability in preserving fine details and achieving high visual fidelity across diverse image categories.

\vspace{2mm}
\subsection{More Generation Results on ImageNet} \label{sec:appendix_generation}
\begin{figure}[htbp]
    \centering
    \includegraphics[width=0.75\linewidth]{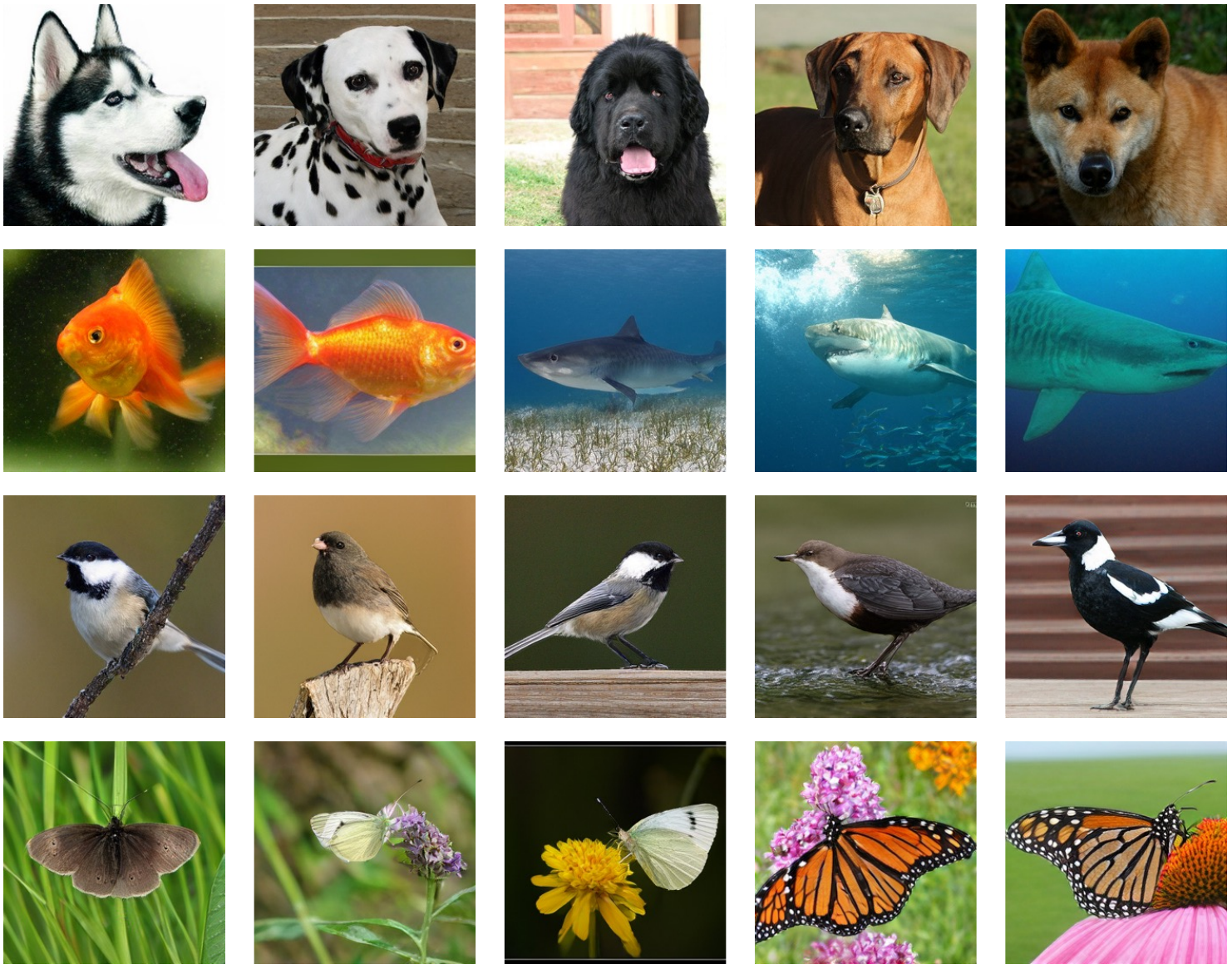}
    \caption{Additional generation results on the ImageNet dataset (SFTok-B).}
    \label{fig:appendix_generation_a}
    \vspace{-4mm}
\end{figure}

\begin{figure}[tbp]
    \centering
    \includegraphics[width=0.75\linewidth]{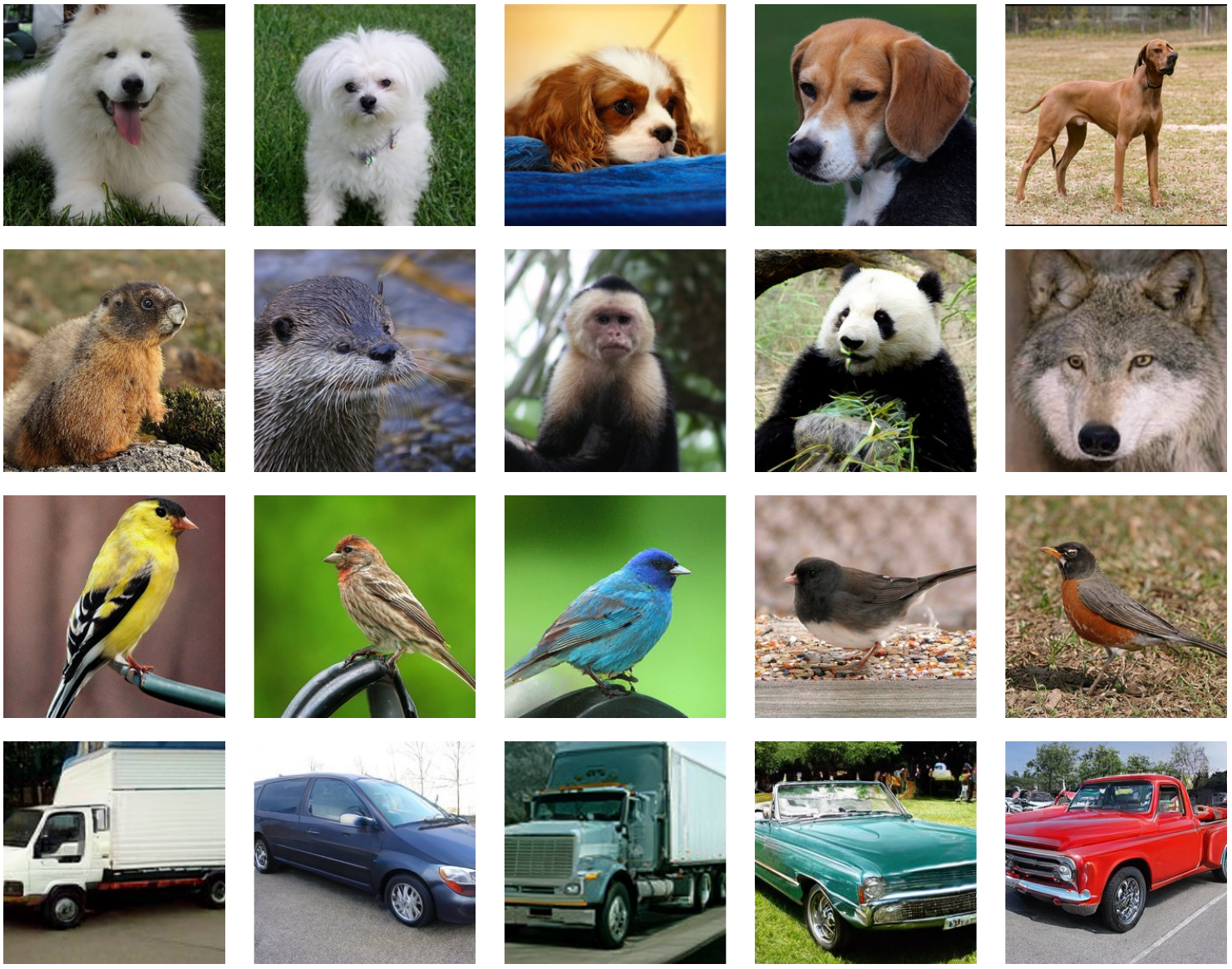}
    \caption{Additional generation results on the ImageNet dataset (SFTok-L).}
    \label{fig:appendix_generation_b}
    \vspace{-4mm}
\end{figure}

This section supplements the main paper with extended image generation results on the ImageNet dataset. 
The samples in \Cref{fig:appendix_generation_a,fig:appendix_generation_b} further demonstrate the capability of our model to produce diverse and high-fidelity images.

\end{document}